%% file: main.tex
\documentclass[10pt,twocolumn,letterpaper]{article}

\usepackage{cvpr}              
\usepackage{graphicx}
\usepackage{amsmath}
\usepackage{amssymb}
\usepackage{booktabs}
\usepackage{bigstrut}
\usepackage{multirow}
\usepackage{array}
\usepackage{tabularx}
\usepackage{tabu}
\usepackage{amsthm}
\usepackage{diagbox}
\usepackage{ulem}
\input{preamble}

\definecolor{cvprblue}{rgb}{0.21,0.49,0.74}
\usepackage[pagebackref,breaklinks,colorlinks,citecolor=cvprblue]{hyperref}

\def\paperID{*****} 
\def\confName{CVPR}
\def\confYear{2024}

\title{Low-Rank Rescaled Vision Transformer Fine-Tuning: A Residual Design Approach}
\author{
	Wei Dong$^{1}$, Xing Zhang$^{2}$, Bihui Chen$^{2}$, Dawei Yan$^{2}$, Zhijun Lin$^{3}$, Qingsen Yan$^{3}$, Peng Wang$^{1}$\thanks{Corresponding author. W. Dong's participation was in part supported by the Natural Science Basic Research Program of Shaanxi (Program No.2024JC-YBMS-464).}, Yang Yang$^{1}$\\
    $^1$School of Computer Science and Engineering, University of Electronic Science and Technology of China\\
    $^2$College of information and control engineering, Xi'an University of Architecture and Technology\\ 
    $^3$School of Computer Science, Northwestern Polytechnical University\\ 
}

\begin{document}
\maketitle
\input{sec/0_abstract}    
\input{sec/1_intro}

\input{sec/2_related_work}

\input{sec/3_method}

\input{sec/4_experiment}

\input{sec/5_conclusion}
{
    \small
    \normalem
    \bibliographystyle{ieeenat_fullname}
    \bibliography{main}
}

\input{sec/X_suppl}

\end{document}

%% file: preamble.tex
\usepackage[dvipsnames]{xcolor}


%% file: sec/0_abstract.tex
\begin{abstract}
Parameter-efficient fine-tuning for pre-trained Vision Transformers aims to adeptly tailor a model to downstream tasks by learning a minimal set of new adaptation parameters while preserving the frozen majority of pre-trained parameters. 
Striking a balance between retaining the generalizable representation capacity of the pre-trained model and acquiring task-specific features poses a key challenge. Currently, there is a lack of focus on guiding this delicate trade-off.
In this study, we approach the problem from the perspective of Singular Value Decomposition (SVD) of pre-trained parameter matrices, providing insights into the tuning dynamics of existing methods.
Building upon this understanding, we propose a Residual-based Low-Rank Rescaling (RLRR) fine-tuning strategy. This strategy not only enhances flexibility in parameter tuning but also ensures that new parameters do not deviate excessively from the pre-trained model through a residual design. Extensive experiments demonstrate that our method achieves competitive performance across various downstream image classification tasks, all while maintaining comparable new parameters.
We believe this work takes a step forward in offering a unified perspective for interpreting existing methods and serves as motivation for the development of new approaches that move closer to effectively considering the crucial trade-off mentioned above. Our code is available at \href{https://github.com/zstarN70/RLRR.git}{https://github.com/zstarN70/RLRR.git}.
\end{abstract}

%% file: sec/1_intro.tex
\section{Introduction}
\label{sec:intro}

In response to the remarkable capabilities demonstrated by large pre-trained models, the paradigm in computer vision and natural language processing has shifted from training task-specific models to fine-tuning a shared pre-trained model~\cite{dosovitskiy2020image,liu2021swin}. Within this trajectory, Parameter-Efficient Fine-Tuning (PEFT) has emerged as an active research area, seeking to adeptly tailor a model to downstream tasks by learning a minimal set of new adaptation parameters while keeping the majority of pre-trained parameters frozen.

The central challenge of PEFT lies in efficiently adapting the pre-trained model to downstream tasks without compromising its generalization capacity. Existing work~\cite{hu2021lora, jia2022visual, houlsby2019parameter, lian2022scaling} has predominantly focused on the efficient adaptation aspect of PEFT, devising various strategies to adjust pre-trained model parameters. 
However, less attention has been given to the crucial task of striking a balance between preserving the pre-trained model's capacity and enabling effective task adaptation.
We believe that the pre-trained model inherently possesses robust generalization capabilities, and the phenomenon of prevalent low-rank strategies~\cite{hu2021lora,lian2022scaling,jie2023fact,dong2023efficient} surpassing full fine-tuning corroborates the existence of significant redundancy within the parameter matrix tuning process. 
In this work, our aim is to take a step forward and explore how to achieve a better trade-off, offering a unified perspective to comprehend this critical balance.

We approach the analysis by viewing each pre-trained parameter matrix through the lens of Singular Value Decomposition (SVD), breaking down the raw matrix into a series of terms. 
Each term is the product of a left-singular column vector, a right-singular row vector, and a corresponding singular value. 
We then examine mainstream PEFT strategies such as adaptation-based methods~\cite{houlsby2019parameter}, LoRA~\cite{hu2021lora}, prompt-tuning~\cite{jia2022visual}, scaling and shifting~\cite{lian2022scaling}, using this framework. This perspective enhances our understanding of these methods, shedding light on how they tune parameters toward downstream tasks and the extent of their tuning.

Building on this analysis, we propose a low-rank rescaled fine-tuning strategy with a residual design. Our fine-tuning is formulated as a combination of a frozen matrix and a low-rank-based rescaling and shifting of the matrix. 
The low-rank rescaling strategy tunes the frozen matrix both row-wise and column-wise, providing enhanced flexibility in matrix tuning. 
The inclusion of the residual term proves crucial in preventing the tuned parameters from deviating excessively from the pre-trained model. 

Extensive experiments demonstrate that our method achieves competitive performance across various downstream image classification tasks while maintaining comparable new parameters. The contributions of this work can be summarized as follows:
\begin{itemize}
	
	\item \textbf{Unified Analytical Framework}: We introduce a unified analytical framework based on SVD to view pre-trained parameter matrices, providing a comprehensive understanding of mainstream PEFT strategies.
	
	\item \textbf{Trade-off Exploration}: Addressing a gap in existing research, we take a significant step forward by exploring the trade-off between preserving the generalization capacity of pre-trained models and efficiently adapting them to downstream tasks in PEFT.
	
	\item \textbf{Proposed Method}: We propose a novel Low-Rank Rescaled Fine-Tuning strategy with a Residual Design. This method formulates fine-tuning as a combination of a frozen matrix and a low-rank-based rescaling and shift, offering enhanced flexibility in matrix tuning.
	
	\item \textbf{Comprehensive Experiments}: Extensive experiments on various downstream image classification tasks showcase the competitiveness of our proposed method, achieving comparable performance with existing strategies while maintaining a minimal set of new parameters.

\end{itemize}

%% file: sec/2_related_work.tex
\section{Related Work}
\label{sec:related_work}

\subsection{Pre-training and Transfer Learning}
Transfer learning, as demonstrated by various studies~\cite{zhuang2020comprehensive,pan2009survey,iman2023review,ying2018transfer}, has proven its adaptability across diverse domains, modalities, and specific task requirements. It has significantly improved performance and convergence speed by pre-training on large-scale datasets and leveraging acquired parameters as initialization for downstream tasks. Large-scale datasets play a pivotal role in this paradigm, contributing to the performance and convergence speed of pre-trained models in downstream tasks. They endow these models with robust generalization capabilities that enhance learning efficiency. Additionally, self-supervised pre-training~\cite{he2022masked,chenempirical} offers further benefits by mitigating costs, time, and quality issues associated with manual data labeling.

In the field of computer vision, earlier studies favor pre-training by the ImageNet-1K dataset~\cite{deng2009imagenet} to attain quicker convergence and enhanced performance in downstream tasks. However, with the advent of larger-scale models like Vision Transformer~\cite{dosovitskiy2020image} (ViT) and Swin Transformer~\cite{liu2021swin}, researchers have shifted toward utilizing more extensive datasets such as ImageNet-21K~\cite{deng2009imagenet} and JFT-300M~\cite{sun2017revisiting}, for pre-training to pursue enhanced training efficiency and robustness. Nevertheless, the adoption of large-scale models presents substantial challenges due to the computational resources required during fine-tuning for downstream tasks. Consequently, researchers have begun exploring methods to achieve efficient fine-tuning.

\subsection{Parameter-Effcient Fine-Tuning}
To mitigate the computational resource challenges posed by exponential parameter growth when fine-tuning the entire network on downstream tasks, PEFT~\cite{jia2022visual,hu2021lora,dong2023efficient,lian2022scaling} endeavors to facilitate the transition of pre-trained models to downstream tasks while significantly reducing the number of trainable parameters compared to full fine-tuning. This reduction aims to minimize training and storage expenses while addressing the risk of overfitting.

In the field of NLP, various PEFT methods have been proposed and have attained significant success~\cite{hu2021lora,lester2021power,liu2023pre, zhang2024composing, huang2023lorahub, luo2023lcm}. Adapter~\cite{houlsby2019parameter}, as one of the primary fine-tuning approaches for large models, introduces a paradigm for fine-tuning through bottleneck structures, entailing the insertion of trainable adapter components into the network structure. Additionally, LoRA~\cite{hu2021lora} employs low-rank decomposition to reduce parameters and treats adapters as side paths to simulate parameter matrix increments during fine-tuning. Subsequently, a multitude of PEFT methods tailored for pre-training ViT models emerged. VPT~\cite{jia2022visual} employs a limited number of trainable parameters in the input and intermediate layers of ViT. It fine-tunes solely these lightweight parameters while maintaining the backbone frozen, resulting in notable performance improvements compared to full fine-tuning.
SSF~\cite{lian2022scaling} introduces a feature modulation method that efficiently transfers features in pre-trained models by scale and shift operations. Unlike sequential adapter insertion approaches, AdaptFormer~\cite{chen2022adaptformer} explores a parallel adapter solution on ViT for various downstream tasks. FacT~\cite{jie2023fact}, based on a tensor decomposition framework, decomposes and reassembles parameter matrices in ViT, allowing lightweight factors to dominate the fine-tuning increment, and only updates the factors during fine-tuning for downstream tasks, resulting in lower fine-tuning costs. ARC~\cite{dong2023efficient} approaches fine-tuning from the perspective of the cross-layer similarity in ViT, using parameter-sharing adapter structures and independent scaling factors, offering a lesser fine-tuning cost than other methods.

\subsection{Discussion to the Proposed Method}
The proposed method incorporates a unique residual structure. Diverging from alternative parallel-structured methods, such as LoRA~\cite{hu2021lora}, which introduces solely low-rank learnable adaptors and can lead to challenges in fine-tuning, our approach navigates the model toward a nuanced balance between optimizing for downstream tasks and preserving the model's intrinsic representational capacity. In contrast to SSF~\cite{lian2022scaling}, we extend our consideration to the adjustment of the singular column vector through a framework rooted in SVD, a dimension that SSF does not encompass. In summary, our study provides a cohesive perspective on past methodologies and presents compelling motivations for this specific strategy.

%% file: sec/3_method.tex
\section{Methodology}
\label{sec:methodology}

In this section, we provide a comprehensive overview of the fundamental concepts related to PEFT methods. We leverage SVD to analyze the pre-trained weight matrices, delving into the underlying mechanisms of popular PEFT approaches within the SVD framework. Our scrutiny is centered on the delicate balance between retaining the generalization capacity of pre-trained parameters and facilitating task-specific adaptation. Concluding this analysis, we introduce our Residual-based Low-Rank Rescaling (RLRR) strategy, designed to optimize this trade-off for enhanced fine-tuning performance.

\subsection{Preliminary Knowledge on PEFT Methods}
ViT is a deep learning model that applies the Transformer~\cite{vaswani2017attention} architecture to computer vision tasks like image classification, originally designed for natural language processing. The ViT model comprises two primary components: a patch embedding layer and a Transformer encoder. The patch embedding layer splits an input image $\mathbf{X}\in \mathbb{R}^{H\times W\times C}$ into a sequence of fixed-size patches, and projects each patch into a high-dimensional vector, \emph{i.e.}, $\mathbf{X}_{\rm patches}\in \mathbb{R}^{N\times (P^{2}\cdot C)}$, where $H$ and $W$ are respectively the height and width of the image resolution $(H, W)$, $(P, P)$ is the resolution of each patch, $C$ is the number of input channels, and $N=H\cdot W/P^{2}$ is the number of tokens. The entire patch embedding layer can be described as follows:
\begin{equation}
	\mathbf{X}_{e}=[\Vec{\boldsymbol{x}}_{\rm cls}^{\top};\mathbf{X}_{\rm patches}\mathbf{W}_{\rm patches}]+\mathbf{X}_{\rm pos},
\end{equation}
where a learnable \emph{class} token $\Vec{\boldsymbol{x}}_{\rm cls} \in \mathbb{R}^{D}$ is concatenated to $\mathbf{X}_{\rm patches}\mathbf{W}_{\rm patches}$ using a linear projection $\mathbf{W}_{\rm patches}\in \mathbb{R}^{(P^{2}\cdot C)\times D}$ and the concatenation operation $[\cdot;\cdot]$. Additionally, position embeddings $\mathbf{X}_{\rm pos} \in \mathbb{R}^{(N+1)\times D}$ are incorporated. The Transformer encoder then processes the patch embeddings using Multi-Head Attention~(MHA) and Feed-Forward Network~(FFN) blocks. In MHA block, Attention Head~(AH) module is defined as:
\begin{equation}
	\begin{split}
		&{\rm AH}_{h}(\mathbf{X}^{(l-1)})= \\
		&{\rm softmax}(\frac{(\mathbf{X}^{(l-1)}\mathbf{W}_{q}^{(l)})(\mathbf{X}^{(l-1)}\mathbf{W}_{k}^{(l)})^{\top}}{\sqrt{D_{h}^{(l)}}})\mathbf{X}^{(l-1)}\mathbf{W}_v^{(l)},
	\end{split}
\end{equation}
\noindent where the weight matrices $\mathbf{W}_{q}^{(l)}\in \mathbb{R}^{D^{(l-1)}\times D_{h}^{(l)}}$, $\mathbf{W}_{k}^{(l)}\in \mathbb{R}^{D^{(l-1)}\times D_{h}^{(l)}}$, and $\mathbf{W}_{v}^{(l)}\in \mathbb{R}^{D^{(l-1)}\times D_{h}^{(l)}}$ are respectively the \emph{query}, \emph{key}, and \emph{value} operations with the feature dimensionality $D_{h}^{(l)}=\frac{D^{(l)}}{M}$ of the output of ${\rm AH}_{h}(\cdot)$ and the number of attention heads $M$. Hence, the whole MHA block is defined as:
\begin{equation}
	\begin{split}
		{\rm MHA}&(\mathbf{X}^{(l-1)})= \\
		&[{\rm AH}_{1}(\mathbf{X}^{(l-1)}), \cdots, {\rm AH}_{M}(\mathbf{X}^{(l-1)})]\mathbf{W}_{o}^{(l)},
	\end{split}
\end{equation}
\noindent with a linear projection $\mathbf{W}_{o}^{(l)}\in \mathbb{R}^{(M\cdot D_{h}^{(l)})\times D^{(l)}}$. We then feed the normalized output $\mathbf{X}^{(l)'}$ of the MHA block into FFN block:
\begin{equation}
	{\rm FFN}(\mathbf{X}^{(l)'})={\rm GELU}(\mathbf{X}^{(l)'}\mathbf{W}_{1}^{(l)})\mathbf{W}_{2}^{(l)},
\end{equation}
\noindent where $\mathbf{W}_{1}^{(l)} \in \mathbb{R}^{D^{(l)}\times 4\cdot D^{(l)}}$ and $\mathbf{W}_{2}^{(l)} \in \mathbb{R}^{4\cdot D^{(l)}\times D^{(l)}}$ denote two linear projection matrices respectively. The whole process of $(l)$-th Transformer encoder layer is defined as:
\begin{equation}
	\begin{split}
		\mathbf{X}^{(l)'}=&{\rm MHA}({\rm LayerNorm}(\mathbf{X}^{(l-1)}))+\mathbf{X}^{(l-1)}, \\
		\mathbf{X}^{(l)}=&{\rm FFN}({\rm LayerNorm}(\mathbf{X}^{(l)'}))+\mathbf{X}^{(l)'}, \\
	\end{split}
\end{equation}
\noindent with ${\rm LayerNorm}(\cdot)$ function to layer representation normalization.

In downstream tasks involving ViT and its variants, three primary types of visual PEFT methods are employed. These methods fine-tune the pre-trained model by utilizing a minimal set of new parameters, and they encompass adaptation-based, prompt-based, and scaling \& shifting-based strategies. More specifically, when considering any weight matrix:
\begin{equation}
	\mathbf{W}^{(l)} \in \{\mathbf{W}_{q}^{(l)}, \mathbf{W}_{k}^{(l)}, \mathbf{W}_{v}^{(l)}, \mathbf{W}_{o}^{(l)}, \mathbf{W}_{1}^{(l)}, \mathbf{W}_{2}^{(l)}\}, 
\end{equation}
the general idea of adaptation-based methods~\cite{houlsby2019parameter} can be defined as Eq.~(\ref{eq:adaptation}) from Table~\ref{table:adaptation_singular}, in which ${\rm Act}(\cdot)$ is the activation function, $\vec{\boldsymbol{b}}^{(l)}$ is the bias weights, and $\mathbf{W}_{\rm down} \in \mathbb{R}^{D^{(l)}\times D^{'}}$ and $\mathbf{W}_{\rm up} \in \mathbb{R}^{D^{'}\times D^{(l)}}$ are down- and up-adapting projection matrices across different layers with the dimensionality $D^{(l)}\gg D^{'}$. A prominent example of an adaptation-based method is Low-Rank Adaptation (LoRA)\cite{hu2021lora}, which can be expressed as Eq.(\ref{eq:LoRA}) in Table~\ref{table:adaptation_singular}.

\begin{table*}[!htbp]
	\centering
	\caption{Examples of PEFT methods and their interpretation under the SVD framework.}
	\renewcommand\arraystretch{2}
	\resizebox{\linewidth}{!}{
		\begin{tabular}{c|c|c}
			\toprule
			\toprule
			\textbf{Visual PEFT Method} & \textbf{Strategy} & \textbf{Spectral Analysis} \\
			
			\hline
			\vspace{3mm}
			 Adaptation-based~\cite{houlsby2019parameter} & \begin{minipage}[c][2.6cm]{.4\textheight}
				\begin{equation}\label{eq:adaptation}
					\mathbf{X}_{\mathrm{FT}}^{(l-1)}=\operatorname{Act}\left(\left(\mathbf{X}^{(l-1)} \mathbf{W}^{(l)}+\vec{\boldsymbol{b}}^{(l) \top}\right) \mathbf{W}_{\mathrm{down}}\right) \mathbf{W}_{\mathrm{up}} ,
			\end{equation}\end{minipage} & \begin{minipage}[c][2.6cm]{.7\textheight}\begin{equation}\label{eq:adaptation_singular}
					\begin{aligned}
						\mathbf{X}_{\mathrm{FT}}^{(l-1)}= & \operatorname{Act}\left(\left(\mathbf{X}^{(l-1)}\left(\lambda_1^{(l)} \vec{\boldsymbol{d}}_1^{(l)} \vec{\boldsymbol{u}}_1^{(l) \top}+\cdots+\lambda_D^{(l)} \vec{\boldsymbol{d}}_D^{(l)} \vec{\boldsymbol{u}}_D^{(l) \top}\right)+\vec{\boldsymbol{b}}^{(l) \top}\right) \mathbf{W}_{\mathrm{down}}\right) \mathbf{W}_{\mathrm{up}} \\
						= & \operatorname{Act}\left(\mathbf{X}^{(l-1)}\left(\lambda_1^{(l)} \vec{\boldsymbol{d}}_1^{(l)} \vec{\boldsymbol{u}}_1^{(l) \top} \mathbf{W}_{\mathrm{down}}+\cdots+\lambda_D^{(l)} \vec{\boldsymbol{d}}_D^{(l)} \vec{\boldsymbol{u}}_D^{(l) \top} \mathbf{W}_{\mathrm{down}}\right)+\vec{\boldsymbol{b}}^{(l) \top} \mathbf{W}_{\text {down }}\right) \mathbf{W}_{\mathrm{up}} \\
						= & \operatorname{Act}\left(\mathbf{X}^{(l-1)}\left(\vec{\boldsymbol{d}}_1^{(l)} \vec{\boldsymbol{u}}_1^{(l) \top} \lambda_1^{(l)} \mathbf{W}_{\mathrm{down}}+\cdots+ \vec{\boldsymbol{d}}_D^{(l)} \vec{\boldsymbol{u}}_D^{(l) \top} \lambda_D^{(l)}\mathbf{W}_{\mathrm{down}}\right)+\vec{\boldsymbol{b}}^{(l) \top} \mathbf{W}_{\text {down }}\right) \mathbf{W}_{\mathrm{up}},
					\end{aligned}
			\end{equation}\end{minipage} \\
			
			\hline
			LoRA adaptation~\cite{hu2021lora}  & \begin{minipage}{.4\textheight}\begin{equation}\label{eq:LoRA}
					\mathbf{X}_{\rm FT}^{(l-1)}=\mathbf{X}^{(l-1)}(\mathbf{W}^{(l)}+\mathbf{W}_{\rm down}\mathbf{W}_{\rm up})+\vec{\boldsymbol{b}}^{(l)\top},
			\end{equation}\end{minipage}     & \begin{minipage}{.7\textheight}\begin{equation}\label{eq:LoRA_singular}
					\begin{aligned}
						\mathbf{X}_{\mathrm{FT}}^{(l-1)}= & \mathbf{X}^{(l-1)}\left(\lambda_1^{(l)} \vec{\boldsymbol{d}}_1^{(l)} \vec{\boldsymbol{u}}_1^{(l) \top}+\cdots+\lambda_D^{(l)} \vec{\boldsymbol{d}}_D^{(l)} \vec{\boldsymbol{u}}_D^{(l) \top}+\mathbf{W}_{\mathrm{down}} \mathbf{W}_{\mathrm{up}}\right)+\vec{\boldsymbol{b}}^{(l) \top},
					\end{aligned}
			\end{equation}\end{minipage} \\
			
			\hline
			Prompt-based~\cite{jia2022visual}   & \begin{minipage}{.4\textheight}\begin{equation}\label{eq:prompt}
					\mathbf{X}_{\mathrm{FT}}^{(l-1)}=\left(\begin{array}{l}
						\mathbf{X}^{(l-1)} \\
						\boldsymbol{\mathbf{\Theta}}^{(l-1)}
					\end{array}\right) \mathbf{W}^{(l)}+\vec{\boldsymbol{b}}^{(l)\top},
			\end{equation}\end{minipage}     & \begin{minipage}{.7\textheight}\begin{equation}\label{eq:prompt_singular}
					\begin{aligned}
						\mathbf{X}_{\mathrm{FT}}^{(l-1)}= & \left(\begin{array}{l}
							\mathbf{X}^{(l-1)} \\
							\mathbf{\Theta}^{(l-1)}
						\end{array}\right)\left(\lambda_1^{(l)} \vec{\boldsymbol{d}}_1^{(l)} \vec{\boldsymbol{u}}_1^{(l) \top}+\cdots+\lambda_D^{(l)} \vec{\boldsymbol{d}}_D^{(l)} \vec{\boldsymbol{u}}_D^{(l) \top}\right)+\vec{\boldsymbol{b}}^{(l) \top} \\
						= & \left(\begin{array}{l}
							\lambda_{1}^{(l)}\mathbf{X}^{(l-1)}\vec{\boldsymbol{d}}_1^{(l)} \vec{\boldsymbol{u}}_1^{(l) \top}+\cdots+\lambda_D^{(l)}\mathbf{X}^{(l-1)}\vec{\boldsymbol{d}}_D^{(l)} \vec{\boldsymbol{u}}_D^{(l) \top} \\
							\lambda_1^{(l)}\mathbf{\Theta}^{(l-1)}\vec{\boldsymbol{d}}_1^{(l)} \vec{\boldsymbol{u}}_1^{(l) \top}+\cdots+\lambda_D^{(l)}\mathbf{\Theta}^{(l-1)}\vec{\boldsymbol{d}}_D^{(l)} \vec{\boldsymbol{u}}_D^{(l) \top}
						\end{array}\right) +\vec{\boldsymbol{b}}^{(l) \top},
					\end{aligned}
			\end{equation}\end{minipage} \\
			
			\hline	
			scaling\&shifting-based~\cite{lian2022scaling}   & \begin{minipage}[c]{.4\textheight}\begin{equation}\label{eq:scaling_shifting}
					\mathbf{X}_{\mathrm{FT}}^{(l-1)}=\left(\mathbf{X} \mathbf{W}^{(l)}+\vec{\boldsymbol{b}}^{(l) \top}\right) \odot \vec{\boldsymbol{s}}^{(l) \top}+\vec{\boldsymbol{f}}^{(l) \top},
			\end{equation}\end{minipage}    & \begin{minipage}[c]{.7\textheight}\begin{equation}\label{eq:scaling_shifting_singular}
					\begin{aligned}
						\mathbf{X}_{\mathrm{FT}}^{(l-1)}= & \left(\mathbf{X} \left(\lambda_1^{(l)} \vec{\boldsymbol{d}}_1^{(l)} \vec{\boldsymbol{u}}_1^{(l) \top}+\cdots+\lambda_D^{(l)} \vec{\boldsymbol{d}}_D^{(l)} \vec{\boldsymbol{u}}_D^{(l) \top}\right)+\vec{\boldsymbol{b}}^{(l) \top}\right) \odot \vec{\boldsymbol{s}}^{(l) \top}+\vec{\boldsymbol{f}}^{(l) \top} \\
						= & \mathbf{X}\left(\lambda_1^{(l)} \vec{\boldsymbol{d}}_1^{(l)} \vec{\boldsymbol{u}}_1^{(l) \top} \odot \vec{\boldsymbol{s}}^{(l) \top}+\cdots+\lambda_D^{(l)} \vec{\boldsymbol{d}}_D^{(l)} \vec{\boldsymbol{u}}_D^{(l) \top} \odot \vec{\boldsymbol{s}}^{(l) \top}\right)+\vec{\boldsymbol{b}}^{(l) \top} \odot \vec{\boldsymbol{s}}^{(l) \top}+\vec{\boldsymbol{f}}^{(l) \top} \\
						= & \mathbf{X}\left( \vec{\boldsymbol{d}}_1^{(l)} \vec{\boldsymbol{u}}_1^{(l) \top} \lambda_1^{(l)}\odot \vec{\boldsymbol{s}}^{(l) \top}+\cdots+ \vec{\boldsymbol{d}}_D^{(l)} \vec{\boldsymbol{u}}_D^{(l) \top}\lambda_D^{(l)} \odot \vec{\boldsymbol{s}}^{(l) \top}\right)+\vec{\boldsymbol{b}}^{(l) \top} \odot \vec{\boldsymbol{s}}^{(l) \top}+\vec{\boldsymbol{f}}^{(l) \top},
					\end{aligned}
			\end{equation}\end{minipage} \\
			\bottomrule
			\bottomrule	
		\end{tabular}
	}
	\label{table:adaptation_singular}
\end{table*}

The second type comprises prompt-based methods~\cite{jia2022visual}, represented by Eq.(\ref{eq:prompt}), where $\mathbf{\Theta}^{(l-1)} \in \mathbb{R}^{T\times D^{(l-1)}}$ constitutes learnable parameters with $T$ virtual tokens. Finally, the third type encompasses scaling \& shifting-based strategies, illustrated by Eq.(\ref{eq:scaling_shifting}), featuring learnable scaling parameters $\vec{\boldsymbol{s}}^{(l)}$, shifting parameters $\vec{\boldsymbol{f}}^{(l)}$, and element-wise Hadamard product denoted by $\odot$.

\subsection{Revisiting Existing PEFT Methods through singular value decomposition}
In this section, we revisit the working mechanisms of existing PEFT methods mentioned above through the lens of SVD. Our goal is to establish a unified framework for understanding the delicate trade-off between retaining the generalization capacity of the pre-trained model and facilitating task-specific adaptation.
We initiate our exploration by employing SVD to decompose the weight matrix $\mathbf{W}^{(l)}$ to:
\begin{equation}
	\mathbf{W}^{(l)}=\lambda_{1}^{(l)}\vec{\boldsymbol{d}}_{1}^{(l)}\vec{\boldsymbol{u}}_{1}^{(l)\top}+\cdots+\lambda_{D}^{(l)}\vec{\boldsymbol{d}}_{D}^{(l)}\vec{\boldsymbol{u}}_{D}^{(l)\top},
\end{equation}
with the spectrum~(\ie singular values) $\{\lambda_{d}^{(l)}\}$, the left singular vector $\vec{\boldsymbol{d}}_{d}^{(l)}$ coming from the left unitary matrix, and the right singular vector $\vec{\boldsymbol{u}}_{d}^{(l)\top}$ coming from the right unitary matrix.  Under this SVD framework, Eqs.~(\ref{eq:adaptation}),~(\ref{eq:LoRA}),~(\ref{eq:prompt}), and~(\ref{eq:scaling_shifting}) can be rewritten as Eqs.~(\ref{eq:adaptation_singular}),~(\ref{eq:LoRA_singular}),~(\ref{eq:prompt_singular}), and~(\ref{eq:scaling_shifting_singular}) in Table~\ref{table:adaptation_singular}.

Upon examining these redefined equations, it becomes evident that general adaptation-based methods involve each singular item under the spectrum, denoted as $\lambda_{d}^{(l)}\vec{\boldsymbol{d}}_{d}^{(l)}\vec{\boldsymbol{u}}_{d}^{(l)\top}\mathbf{W}_{\rm down}$. The down-adapting projection matrix $\mathbf{W}_{\rm down}$ is directly applied to the right singular vector $\vec{\boldsymbol{u}}_{d}^{(l)}$. However, this direct application compromises the spatial structure, including the orthogonality of these right singular matrix $[{\vec{\boldsymbol{u}}_{1}^{(l)}}, {\vec{\boldsymbol{u}}_{2}^{(l)}},\ldots, {\vec{\boldsymbol{u}}_{d}^{(l)}}]^{\top}$, thereby affecting the representation capacity of the pre-trained model.
Similarly, in prompt-based methods, the learnable tokens $\mathbf{\Theta}$ directly interat with the left singular vector $\vec{\boldsymbol{d}}_{d}^{(l)}$ in Eq.~(\ref{eq:prompt_singular}). However, this direct interaction has the potential to excessively influence the tuning, deviating significantly from the pre-trained model.
Scaling\&shifting-based methods has the same defect due to the element-wise multiplication $\vec{\boldsymbol{u}}_d^{(l) \top} \odot \vec{\boldsymbol{s}}^{(l) \top}$ in Eq.~(\ref{eq:scaling_shifting_singular}). Additionally, over-adaptation may perturb the spectrum, affecting one side of the weight capacity. Specifically, $\lambda_d^{(l)}\mathbf{W}_{\mathrm{down}}$ in Eq.~(\ref{eq:adaptation_singular}), $\lambda_d^{(l)}\mathbf{\Theta}^{(l-1)}$ in Eq.~(\ref{eq:prompt_singular}), and $\lambda_d^{(l)} \odot \vec{\boldsymbol{s}}^{(l) \top}$ in Eq.~(\ref{eq:scaling_shifting_singular}) demonstrate the impact to the singular spectrum. Improper initialization of parameters, such as $\mathbf{W}_{\mathrm{down}}$, $\mathbf{\Theta}^{(l-1)}$, and $\vec{\boldsymbol{s}}^{(l)}$, can lead to spectrum distortion and the loss of the original weight capacity.

In contrast, from Eq.~(\ref{eq:LoRA_singular}), we observe that the sole low-rank adaption item $\mathbf{W}_{\rm down}\mathbf{W}_{\rm up}$ of LoRA method adapts weakly to each of all singular items $\{\lambda_{d}^{(l)}\vec{\boldsymbol{d}}_{d}^{(l)}\vec{\boldsymbol{u}}_{d}^{(l)\top}\}$ when the dimensionality $D$ of the weight matrix $\mathbf{W}^{(l)}$ is large. This slight perturbation may marginally change the weight spectrum and singular vectors in which the representation capacity of the pre-trained model can not be smoothly adapted to downstream tasks.

\begin{figure*}[!htbp]
	\centering
	\includegraphics[width=1.0\linewidth]{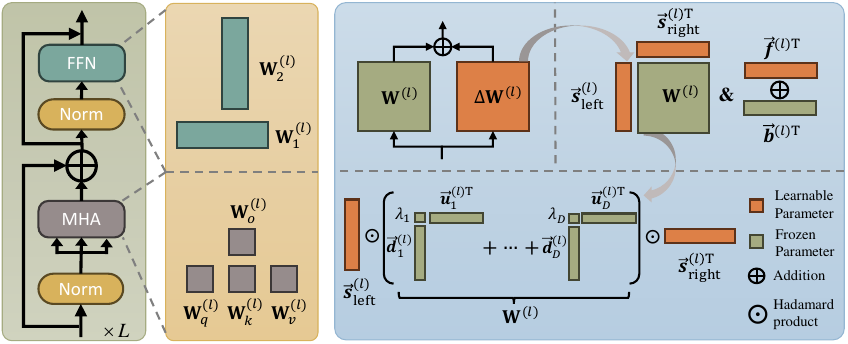}
	\caption{Illustration of the proposed RLRR method. For any weight matrix $\mathbf{W}^{(l)}$ in the MHA and FFN modules, we fine-tune the frozen pre-training parameter matrix using a residual structure. This involves combining the frozen matrix with a low-rank-based scaling and shifting operation \emph{i.e.}, $\bigtriangleup\mathbf{W}^{(l)}$. From the perspective of SVD, scaling vectors $\vec{\boldsymbol{s}}_{\rm left}^{(l)}$ and $\vec{\boldsymbol{s}}_{\rm right}^{(l)}$ and shifting vector $\vec{\boldsymbol{f}}_{(l)}$ can also be interpreted as adjustments to the rows and columns of the pre-training matrix $\mathbf{W}^{(l)}$.}
	\label{figMethod}
	\vspace{-1.5em}
\end{figure*}

\subsection{Residual-based Low-Rank Rescaling~(RLRR) Method}
To balance the trade-off between over-adaptation and under-adaptation in downstream tasks, we propose a simple yet effective method, namely, the RLRR strategy as shown in Fig.~\ref{figMethod}. It can be derived from the aforementioned unified framework:
\begin{equation}\label{eq:RLRR}
	\begin{split}
		\mathbf{X}_{\rm FT}^{(l-1)}=&\mathbf{X}^{(l-1)}(\mathbf{W}^{(l)}+\bigtriangleup\mathbf{W}^{(l)})+\vec{\boldsymbol{b}}^{(l)\top}+\vec{\boldsymbol{f}}^{(l)\top} \\
		=&\mathbf{X}^{(l-1)}(\mathbf{W}^{(l)}+\vec{\boldsymbol{s}}_{\rm left}^{(l)}\odot\mathbf{W}^{(l)}\odot\vec{\boldsymbol{s}}_{\rm right}^{(l)\top}) \\
		&+\vec{\boldsymbol{b}}^{(l)\top}+\vec{\boldsymbol{f}}^{(l)\top},
	\end{split}
\end{equation}
in which we add scales $\vec{\boldsymbol{s}}_{\rm left}^{(l)}$ and $\vec{\boldsymbol{s}}_{\rm right}^{(l)}$ to both side of weight matrix $\mathbf{W}^{(l)}$, making it more flexible compared to SSF~\cite{lian2022scaling} when learning the features of downstream tasks.

In Eq.~(\ref{eq:RLRR}), we also add the frozen weights $\mathbf{W}^{(l)}$ to the fine-tuning item $\bigtriangleup\mathbf{W}^{(l)}=\vec{\boldsymbol{s}}_{\rm left}^{(l)}\odot\mathbf{W}^{(l)}\odot\vec{\boldsymbol{s}}_{\rm right}^{(l)\top}$ with learnable parameters $\vec{\boldsymbol{s}}_{\rm left}^{(l)}$, $\vec{\boldsymbol{s}}_{\rm right}^{(l)}$, and $\vec{\boldsymbol{f}}^{(l)}$. By doing this, RLRR strategy can trade off the over- and under-adaption. Concretely, we expand Eq.~(\ref{eq:RLRR}) to:
\begin{equation}\label{eq:RLRR_extension}
	\begin{split}
		\mathbf{X}_{\rm FT}^{(l-1)}=&\mathbf{X}^{(l-1)}(\lambda_{1}^{(l)}\vec{\boldsymbol{d}}_{1}^{(l)}\vec{\boldsymbol{u}}_{1}^{(l)\top}+\cdots+\lambda_{D}^{(l)}\vec{\boldsymbol{d}}_{D}^{(l)}\vec{\boldsymbol{u}}_{D}^{(l)\top} \\
		&+\vec{\boldsymbol{s}}_{\rm left}^{(l)}\odot(\lambda_{1}^{(l)}\vec{\boldsymbol{d}}_{1}^{(l)}\vec{\boldsymbol{u}}_{1}^{(l)\top}+\cdots \\
		&+\lambda_{D}^{(l)}\vec{\boldsymbol{d}}_{D}^{(l)}\vec{\boldsymbol{u}}_{D}^{(l)\top})\odot\vec{\boldsymbol{s}}_{\rm right}^{(l)\top})+\vec{\boldsymbol{b}}^{(l)\top}+\vec{\boldsymbol{f}}^{(l)\top},
	\end{split}
\end{equation}
in which we get the singular item: 
\begin{equation}\label{eq:RLRR_singular_item}
	\begin{split}
		&\mathbf{W}_{\rm item}= \\
		&\lambda_{d}^{(l)}\vec{\boldsymbol{d}}_{d}^{(l)}\vec{\boldsymbol{u}}_{d}^{(l)\top}+\lambda_{d}^{(l)}\vec{\boldsymbol{s}}_{\rm left}^{(l)}\odot\vec{\boldsymbol{d}}_{d}^{(l)}\vec{\boldsymbol{u}}_{d}^{(l)\top}\odot\vec{\boldsymbol{s}}_{\rm right}^{(l)\top},
	\end{split}
\end{equation}
\noindent and each element therein is:
\begin{equation}\label{eq:RLRR_singular_item_each}
	\begin{split}
		&\mathbf{W}_{{\rm item}[i, j]}=\lambda_{d}^{(l)}\vec{\boldsymbol{d}}_{d[i]}^{(l)}\vec{\boldsymbol{u}}_{d[j]}^{(l)}+\lambda_{d}^{(l)}\vec{\boldsymbol{s}}_{{\rm left}[i]}^{(l)}\vec{\boldsymbol{d}}_{d[i]}^{(l)}\vec{\boldsymbol{u}}_{d[j]}^{(l)}\vec{\boldsymbol{s}}_{{\rm right}[j]}^{(l)} \\
		&=(1+\vec{\boldsymbol{s}}_{{\rm left}[i]}^{(l)}\vec{\boldsymbol{s}}_{{\rm right}[j]}^{(l)})\lambda_{d}^{(l)}\vec{\boldsymbol{d}}_{d[i]}^{(l)}\vec{\boldsymbol{u}}_{d[j]}^{(l)}.
	\end{split}
\end{equation}
There is a constant term $1$ in Eq.~(\ref{eq:RLRR_singular_item_each}) that can fix the intrinsical representation capacity to the pre-trained model and meanwhile leverage the fine-tuning item $\vec{\boldsymbol{s}}_{{\rm left}[i]}^{(l)}\vec{\boldsymbol{s}}_{{\rm right}[j]}^{(l)}$ to adaptively adjust such model capacity to learn the downstream tasks.

\noindent 
\textbf{Re-parameterization}.
Similar to previous methods~\cite{lian2022scaling}, our adjustments to the parameter matrices are linear operations. This allows us to seamlessly absorb the scaling and shifting operations into the original parameter matrices by re-parameterizing as follows:
\begin{equation}\label{eq:re_parameter}
	\begin{aligned}
		\mathbf{W}_{\rm re-param}^{(l)}&=\mathbf{W}^{(l)}+\Delta{\mathbf{W}^{(l)}} \\ &=\left(\mathbf{1}+\vec{\boldsymbol{s}}_{\rm left}^{(l)}\vec{\boldsymbol{s}}_{\rm right}^{(l)\top}\right)\odot \mathbf{W}^{(l)} ,\\ 
		\vec{\boldsymbol{b}}_{\rm re-param}^{(l)}&=\vec{\boldsymbol{b}}^{(l)}+\vec{\boldsymbol{f}}^{(l)} ,
	\end{aligned}
\end{equation}
where $\mathbf{1}$ denotes a matrix involving all elements to $1$, with its dimensions consistent with the original parameter matrix $\mathbf{W}^{(l)}$ in the $(l)$-th layer. The vectors $\vec{\boldsymbol{s}}_{\rm left}^{(l)}$ and $\vec{\boldsymbol{s}}_{\rm right}^{(l)}$ denote the scaling parameters and the shifting parameters is the $\vec{\boldsymbol{f}}$ vector. Eq.~(\ref{eq:re_parameter}) implies that we can merge $\vec{\boldsymbol{s}}_{\rm left}^{(l)}$, $\vec{\boldsymbol{s}}_{\rm right}^{(l)}$, and $\vec{\boldsymbol{f}}^{(l)}$ into the original parameter matrix $\mathbf{W}^{(l)}$ by linear operations without requiring extra storage space during inference.

%% file: sec/4_experiment.tex
\section{Experiments}
\noindent
\label{sec:exp}
\subsection{Experimental Settings}

\begin{table*}[!htbp]
	\centering
	\caption{Performance comparison of RLRR with the baseline and state-of-the-art efficient adaptive methods on the VTAB-1k benchmark. All methods leverage ViT-B/16 pre-trained on ImageNet-21k as the backbone. Furthermore, SSF, ARC*, and RLRR* utilize the augmented ViT backbone by AugReg~\cite{steiner2021train}. Bold font denotes state-of-the-art performance, while underlined results indicate sub-optimal performance. }
	\resizebox{\linewidth}{!}{
		\begin{tabular}{c|ccccccc|c|cccc|c|cccccccc|c|cc}
			\toprule
			\toprule
			\multirow{2}[2]{*}{\diagbox{\textbf{Methods}}{\textbf{\rotatebox{0}{Datasets}}}} & \multicolumn{8}{c|}{\textbf{Natural}}                         & \multicolumn{5}{c|}{\textbf{Specialized}} & \multicolumn{9}{c|}{\textbf{Structed}}                                &       &  \\
			& \rotatebox{90}{\textbf{CIFAR-100}} & \rotatebox{90}{\textbf{Caltech101}} & \rotatebox{90}{\textbf{DTD}} & \rotatebox{90}{\textbf{Flowers102}} & \rotatebox{90}{\textbf{Pets}} & \rotatebox{90}{\textbf{SVNH}} & \multicolumn{1}{c}{\rotatebox{90}{\textbf{Sun397}}} & \rotatebox{90}{\textbf{Mean}} & \rotatebox{90}{\textbf{Camelyon}} & \rotatebox{90}{\textbf{EuroSAT}} & \rotatebox{90}{\textbf{Resisc45}} & \multicolumn{1}{c}{\rotatebox{90}{\textbf{Retinopathy}}} & \rotatebox{90}{\textbf{Mean}} & \rotatebox{90}{\textbf{Clevr-Count}} & \rotatebox{90}{\textbf{Clevr-Dist}} & \rotatebox{90}{\textbf{DMLab}} & \rotatebox{90}{\textbf{KITTI-Dist}} & \rotatebox{90}{\textbf{dSpr-Loc}} & \rotatebox{90}{\textbf{dSpr-Ori}} & \rotatebox{90}{\textbf{sNORB-Azim}} & \multicolumn{1}{c}{\rotatebox{90}{\textbf{sNORB-Ele}}} & \rotatebox{90}{\textbf{Mean}} & \rotatebox{90}{\textbf{Mean Total}} & \rotatebox{90}{\textbf{Params.(M)}} \\
			\midrule
			\midrule
			Full fine-tuning & 68.9  & 87.7  & 64.3  & 97.2  & 86.9  & 87.4  & 38.8  & 75.9  & 79.7  & 95.7  & 84.2  & 73.9  & 83.4  & 56.3  & 58.6  & 41.7  & 65.5  & 57.5  & 46.7  & 25.7  & 29.1  & 47.6  & 65.6  & 85.80  \\
			Linear probing & 63.4  & 85.0  & 63.2  & 97.0  & 86.3  & 36.6  & 51.0  & 68.9  & 78.5  & 87.5  & 68.6  & 74.0  & 77.2  & 34.3  & 30.6  & 33.2  & 55.4  & 12.5  & 20.0  & 9.6   & 19.2  & 26.9  & 52.9  & 0.04  \\
			\hline
			Adapter~\cite{houlsby2019parameter} & 74.1  & 86.1  & 63.2  & 97.7  & 87.0  & 34.6  & 50.8  & 70.5  & 76.3  & 88.0  & 73.1  & 70.5  & 77.0  & 45.7  & 37.4  & 31.2  & 53.2  & 30.3  & 25.4  & 13.8  & 22.1  & 32.4  & 55.8  & 0.27  \\
			Bias~\cite{zaken2022bitfit}  & 72.8  & 87.0  & 59.2  & 97.5  & 85.3  & 59.9  & 51.4  & 73.3  & 78.7  & 91.6  & 72.9  & 69.8  & 78.3  & 61.5  & 55.6  & 32.4  & 55.9  & 66.6  & 40.0  & 15.7  & 25.1  & 44.1  & 62.1  & 0.14  \\
			VPT-Shallow~\cite{jia2022visual} & \uline{77.7} & 86.9  & 62.6  & 97.5  & 87.3  & 74.5  & 51.2  & 76.8  & 78.2  & 92.0  & 75.6  & 72.9  & 79.7  & 50.5  & 58.6  & 40.5  & 67.1  & 68.7  & 36.1  & 20.2  & 34.1  & 47.0  & 64.9  & 0.11  \\
			VPT-Deep~\cite{jia2022visual} & \textbf{78.8} & 90.8  & 65.8  & 98.0  & 88.3  & 78.1  & 49.6  & 78.5  & 81.8  & \uline{96.1}  & 83.4  & 68.4  & 82.4  & 68.5  & 60.0  & 46.5  & 72.8  & 73.6  & 47.9  & 32.9  & 37.8  & 55.0  & 69.4  & 0.60  \\
			LORA~\cite{hu2021lora}  & 67.1  & \uline{91.4} & 69.4  & 98.8  & 90.4  & 85.3  & 54.0  & 79.5  & \uline{84.9} & 95.3  & 84.4  & 73.6  & 84.6  & \textbf{82.9} & \textbf{69.2} & \uline{49.8} & 78.5  & 75.7  & 47.1  & 31.0  & \textbf{44.0} & 59.8  & 72.3  & 0.29  \\
			AdaptFormer~\cite{chen2022adaptformer} & 70.8  & 91.2  & 70.5  & \uline{99.1} & 90.9  & 86.6  & \uline{54.8}  & 80.6  & 83.0  & 95.8  & 84.4  & \textbf{76.3} & 84.9  & 81.9 & 64.3  & 49.3  & 80.3  & 76.3  & 45.7  & 31.7  & 41.1  & 58.8  & 72.3  & 0.16  \\
			FacT-TK$_{\leq32}$\cite{jie2023fact} & 70.6  & 90.6  & 70.8  & \uline{99.1} & 90.7  & 88.6  & 54.1  & 80.6  & 84.8  & \textbf{96.2} & 84.5  & 75.7  & \uline{85.3}  & \uline{82.6}  & \uline{68.2} & \uline{49.8} & 80.7  & \uline{80.8} & 47.4  & \uline{33.2} & 43.0  & \uline{60.7} & 73.2  & 0.07  \\
			ARC~\cite{dong2023efficient}   & 72.2  & 90.1  & \uline{72.7} & 99.0  & \uline{91.0} & \textbf{91.9} & 54.4 & \uline{81.6} & \uline{84.9} & 95.7 & \textbf{86.7} & 75.8  & \textbf{85.8} & 80.7  & 67.1  & 48.7  & \uline{81.6} & 79.2  & \uline{51.0} & 31.4  & 39.9  & 60.0  & \uline{73.4}  & 0.13  \\
			RLRR   & 75.6  & \textbf{92.4} & \textbf{72.9} & \textbf{99.3} & \textbf{91.5} & \uline{89.8} & \textbf{57.0} & \textbf{82.7} & \textbf{86.8} & 95.2  & \uline{85.3} & \uline{75.9} & \textbf{85.8} & 79.7  & 64.2  & \textbf{53.9} & \textbf{82.1} & \textbf{83.9} & \textbf{53.7} & \textbf{33.4} & \uline{43.6} & \textbf{61.8} & \textbf{74.5} & 0.33  \\
			\hline
			SSF~\cite{lian2022scaling}   & 69.0  & \uline{92.6} & 75.1  & 99.4  & 91.8  & 90.2  & \uline{52.9} & 81.6  & \uline{87.4} & 95.9  & 87.4  & 75.5  & 86.6  & 75.9  & \uline{62.3} & 53.3  & 80.6  & 77.3  & \uline{54.9} & 29.5  & 37.9  & 59.0  & 73.1  & 0.24  \\
			ARC*~\cite{dong2023efficient}  & \uline{71.2} & 90.9  & \uline{75.9} & \uline{99.5} & \uline{92.1} & \uline{90.8} & 52.0  & \uline{81.8} & \uline{87.4} & \textbf{96.5} & \uline{87.6} & \textbf{76.4} & \uline{87.0} & \textbf{83.3} & 61.1  & \textbf{54.6} & \uline{81.7} & \uline{81.0} & \textbf{57.0} & \uline{30.9} & \uline{41.3} & \uline{61.4} & \uline{74.3} & 0.13  \\
			RLRR*  & \textbf{76.7} & \textbf{92.7} & \textbf{76.3} & \textbf{99.6} & \textbf{92.6} & \textbf{91.8} & \textbf{56.0} & \textbf{83.7} & \textbf{87.8} & \uline{96.2} & \textbf{89.1} & \uline{76.3} & \textbf{87.3} & \uline{80.4} & \textbf{63.3} & \uline{54.5} & \textbf{83.3} & \textbf{83.0} & 53.7  & \textbf{32.0} & \textbf{41.7} & \textbf{61.5} & \textbf{75.1} & 0.33  \\
			\bottomrule
			\bottomrule
		\end{tabular}
	}
	\label{vitb_vtab}
	\vspace{-1.em}
\end{table*}

\textbf{Downstream Tasks}.
Following the previous works~\cite{jia2022visual, lian2022scaling, dong2023efficient}, we evaluate RLRR on a collection of five Fine-Grained Visual Classification (FGVC) datasets and the VTAB-1k benchmark. FGVC consists of \textit{CUB-200-2011}~\cite{wah2011caltech}, \textit{NABirds}~\cite{van2015building}, \textit{Oxford Flowers}~\cite{nilsback2008automated}, \textit{Stanford Dogs}~\cite{khosla2011novel}, and \textit{Stanford Cars}~\cite{gebru2017fine}. We follow the data partitioning scheme established in VPT~\cite{jia2022visual} to maintain consistency. VTAB-1k~\cite{zhai2019large} is a benchmark that contains 19 diverse visual classification tasks, which are divided into three groups: \textit{Natural}, \textit{Specialized}, and \textit{Structured}. \textit{Natural} group corresponds to images from daily life, \textit{Specialized} group includes medical and remote sensing images captured by specialized devices, and \textit{Structured} group contains synthetic images from simulated environments. Each task contains only 1000 images for training, covering various potential downstream tasks such as classification, object counting, and depth estimation. Consequently, it serves as a comprehensive measurement for evaluating the efficacy of fine-tuning methodologies.

\noindent
\textbf{Pre-trained Backbones}. 
We employ ViT~\cite{dosovitskiy2020image} and Swin Transformer~\cite{liu2021swin} as backbones to evaluate our approach. Furthermore, we employ three different variants of ViT (\emph{i.e.} ViT-Base, ViT-Large, ViT-Huge) to demonstrate the versatility of RLRR. All of these backbone architectures leverage parameters pre-trained on the ImageNet21K dataset~\cite{deng2009imagenet}, preserving the default configurations, which include the number of image patches and the dimensions of the features in the hidden layers. Moreover, we note that the SSF~\cite{lian2022scaling} employs a ViT backbone that is augmented with AugReg~\cite{steiner2021train}. To guarantee a fair comparison, we have carried out independent experiments with this augmentation strategy, as presented in Table~\ref{vitb_vtab} and Table~\ref{vitb_fgvc}.

\noindent
\textbf{Baselines and Existing PEFT methods}.
We evaluate the performance of RLRR by comparing it with two baseline methods and several well-known PEFT approaches including Adapter~\cite{houlsby2019parameter}, Bias~\cite{zaken2022bitfit}, LoRA~\cite{hu2021lora}, VPT~\cite{jia2022visual}, AdaptFormer~\cite{chen2022adaptformer}, FacT~\cite{jie2023fact} and ARC~\cite{dong2023efficient}. The two baseline methods are (1) Full Fine-tuning, which updates all parameters of the pre-trained model using the training data from the downstream task, and (2) Linear Probing, which involves training only the linear classification head for the downstream task while keeping the rest of the pre-trained parameters frozen.

\noindent
\textbf{Implementation Details}.
In this work, we implement standard data augmentation following VPT~\cite{jia2022visual} during the training phase. For five FGVC datasets, we apply random horizontal flips and randomly resize crop to $224 \times 224$ pixels. For the VTAB-1k benchmark, images are resized to $224 \times 224$ pixels, and we employ random horizontal flips on the 19 datasets. We conduct a grid search to optimize hyper-parameters specific to tuning, such as learning rate and weight decay. All experiments are conducted using the PyTorch framework~\cite{paszke2019pytorch} on an NVIDIA A800 GPU with 80 GB of memory.

\begin{table*}[!htbp] \scriptsize
	\centering
	\caption{Performance comparison of RLRR with baseline and state-of-the-art PEFT methods on five FGVC datasets. All experiments use ViT-B/16 pretrained on ImageNet-21k as the backbone. SSF, ARC*, and RLRR* leverage the augmented ViT backbone by AugReg~\cite{steiner2021train}.}
		\begin{tabular}{c|c|c|c|c|c|cc}
			\toprule
			\toprule
			\diagbox{\textbf{Methods}}{\textbf{Datasets}} & \multicolumn{1}{c|}{\textbf{CUB-200-2011}} & \textbf{NABirds} & \multicolumn{1}{c|}{\textbf{Oxford Flowers}} & \multicolumn{1}{c|}{\textbf{Stanford Dogs}} & \multicolumn{1}{c|}{\textbf{Stanford Cars}} & \textbf{Mean Total} & \multicolumn{1}{c}{\textbf{Params. (M)}} \\
			\midrule
			\midrule
			Full fine-tuning & 87.3  & 82.7  & 98.8  & 89.4  & 84.5  & 88.5  & 85.98  \\
			Linear probing & 85.3  & 75.9  & 97.9  & 86.2  & 51.3  & 79.3  & 0.18  \\
			\hline
			Adapter~\cite{houlsby2019parameter} & 87.1  & 84.3  & 98.5  & 89.8  & 68.6  & 85.7  & 0.41  \\
			Bias~\cite{zaken2022bitfit}  & 88.4  & 84.2  & 98.8  & 91.2  & 79.4  & 88.4  & 0.28  \\
			VPT-Shallow~\cite{jia2022visual} & 86.7  & 78.8  & 98.4  & 90.7  & 68.7  & 84.6  & 0.25  \\
			VPT-Deep~\cite{jia2022visual} & \uline{88.5}  & 84.2  & 99.0  & 90.2  & 83.6  & 89.1  & 0.85  \\
			LoRA~\cite{hu2021lora}  & 88.3  & \textbf{85.6} & 99.2  & 91.0  & 83.2  & 89.5  & 0.44  \\
			ARC~\cite{dong2023efficient}   & \uline{88.5} & \uline{85.3} & \uline{99.3} & \uline{91.9} & \uline{85.7} & \uline{90.1} & 0.25  \\
			RLRR   & \textbf{89.3} & 84.7  & \textbf{99.5} & \textbf{92.0} & \textbf{87.0} & \textbf{90.4} & 0.47  \\
			\hline
			SSF~\cite{lian2022scaling}   & \uline{89.5} & \textbf{85.7} & \uline{99.6} & \uline{89.6} & 89.2  & \uline{90.7} & 0.39  \\
			ARC*~\cite{dong2023efficient}  & 89.3  & \textbf{85.7} & \textbf{99.7} & 89.1  & \uline{89.5} & \uline{90.7} & 0.25  \\
			RLRR*  & \textbf{89.8} & \uline{85.3} & \uline{99.6} & \textbf{90.0} & \textbf{90.4} & \textbf{91.0} & 0.47  \\
			\bottomrule
			\bottomrule
		\end{tabular}%
	\label{vitb_fgvc}%
\end{table*}%

\begin{table*}[!htbp]
	\centering
	\caption{Performance comparison on VTAB-1k using VIT-Large and VIT-Huge pre-trained on ImageNet-21k as the backbone. "$(\cdot)$" indicates the number of tasks in the subgroup. Detailed results are presented in the Appendix.}
	\resizebox{\linewidth}{!}{
		\begin{tabular}{c|ccc|cc|ccc|cc}
			\toprule
			\toprule
			\multicolumn{1}{c|}{\multirow{2}[2]{*}{\diagbox{\textbf{Methods}}{\textbf{Datasets}}}} & \multicolumn{5}{c|}{(a) ViT-Large} & \multicolumn{5}{c}{(b) ViT-Huge} \\
			& \textbf{Natural (7)} & \textbf{Specialized (4)} & \multicolumn{1}{c}{\textbf{Structed (8)}} & \textbf{Mean} & \textbf{Params.(M)} & \textbf{Natural (7)} & \textbf{Specialized (4)} & \multicolumn{1}{c}{\textbf{Structed (8)}} & \textbf{Mean} & \textbf{Params.(M)} \\
			\midrule
			\midrule
			Full fine-tuning & 74.7  & 83.8  & 48.1  & 65.4  & 303.40  & 70.9  & 83.6  & 46.0  & 63.1  & 630.90  \\
			Linear probing & 70.9  & 69.1  & 25.8  & 51.5  & 0.05  & 67.9  & 79.0  & 26.1  & 52.7  & 0.06  \\
			\hline
			Adapter~\cite{houlsby2019parameter} & 68.6  & 73.5  & 29.0  & 52.9  & 2.38  & 68.1  & 76.4  & 24.5  & 51.5  & 5.78  \\
			Bias~\cite{zaken2022bitfit}  & 70.5  & 73.8  & 41.2  & 58.9  & 0.32  & 70.3  & 78.9  & 41.7  & 60.1  & 0.52  \\
			VPT-Shallow~\cite{jia2022visual} & 78.7  & 79.9  & 40.6  & 62.9  & 0.15  & 74.8  & 81.2  & 43.0  & 62.8  & 0.18  \\
			VPT-Deep~\cite{jia2022visual} & \uline{82.5}  & 83.9  & 54.1  & 70.8  & 0.49  & 77.9  & 83.3  & 52.2  & 68.2  & 0.96  \\
			LoRA~\cite{hu2021lora}  & 81.4  & 85.0  & 57.3  & 72.0  & 0.74  & 77.1  & 83.5  & 55.4  & 69.3  & 1.21  \\
			SSF~\cite{lian2022scaling}  & 81.9  & 85.2  & \uline{59.0}  & \uline{73.0}  & 0.60  & 79.0  & 83.1  & \uline{56.6}  & \uline{70.4}  &  0.97 \\
			ARC~\cite{dong2023efficient}   & 82.3  & \uline{85.6}  & 57.3  & 72.5  & 0.18  & \uline{79.1}  & \uline{84.8}  & 53.7  & 69.6  & 0.22  \\
			\hline
			RLRR   & \textbf{83.9} & \textbf{86.4} & \textbf{61.9} & \textbf{75.2} & 0.82  & \textbf{79.4} & \textbf{85.1} & \textbf{59.0} & \textbf{72.0} & 1.33 \\
			\bottomrule
			\bottomrule
		\end{tabular}%
	}
	\label{vitlhs_vtab}
	\vspace{-1.em}
\end{table*}

\subsection{Experimental Comparisons}
In this section, we conduct a comprehensive comparison of our RLRR method with baseline models and other state-of-the-art approaches using two sets of visual adaptation benchmarks. We evaluate the classification accuracy of each method across a range of downstream tasks and examine the number of trainable parameters during the fine-tuning phase. The outcomes of these evaluations are detailed in Table~\ref{vitb_vtab} and Table~\ref{vitb_fgvc}. Based on the findings, we make the following observations:

(1)~RLRR approach yields results that are competitive with both baseline methods and prior state-of-the-art PEFT methods. Notably, RLRR attains superior performance on the majority of datasets across two visual adaptation benchmarks, outperforming most existing fine-tuning approaches. It also maintains a competitive number of trainable parameters, suggesting that RLRR achieves high efficiency without incurring excessive computational costs. In particular, on the VTAB-1k benchmark, our method excels in more than half of the 19 datasets, achieving a 1.1\% improvement (74.5\% $vs.$ 73.4\%) over the plain pre-trained model and a 0.8\% increase (75.1\% $vs.$ 74.3\%) over the AugReg-enhanced model relative to the latest PEFT methods. Moreover, RLRR demonstrates optimal performance in 7 out of 10 assessments on the FGVC dataset using two versions of pre-trained model, underscoring its consistent adaptability and robustness across varied downstream tasks.

Additionally, it is noteworthy that RLRR and LoRA have comparable parameter counts. However, RLRR significantly outperforms LoRA in downstream tasks. This underscores the superior design of RLRR, which leverages the pre-trained parameter matrix as the foundation for the residual term, thus preventing the potential pitfalls of over or under adaptation in downstream tasks. The tuning of the residual term also benefits from the high-efficiency parameter adjustment inherent in the well-structured design of SSF. Furthermore, our approach incorporates a rescaling weight matrix design, which provides greater flexibility than that of SSF. In contrast to VPT, our method obviates the need for designing complex, task-specific trainable parameters or for intricate injection selections within the partial modules of ViT, thereby avoiding additional computational overhead.

(2)~In contrast to PEFT solutions, Full fine-tuning does not yield significant improvements. In fact, performance can decline even with the increase in the number of updated parameters. We attribute this to the loss of the generalization ability of the pre-trained model, which was acquired from large-scale datasets, leading to overfitting on the training set for downstream tasks. In practice, as a commonly adopted strategy in transfer learning, full fine-tuning necessitates extensive data and meticulous experimental setups to prevent overfitting. Especially on the VTAB-1k benchmark with only 1000 images for training, besides fine-tuning the entire model, numerous adaptation methods often find themselves in the dilemma of overfitting. This underscores the effectiveness and promise of lightweight adaptation designs.

\noindent
\textbf{Experiments on larger-scale ViT backbones. }
Beyond the commonly employed ViT-B/16 backbone for evaluations, we expand our experiments to include larger backbones, ViT-L/16 and ViT-H/14, to verify the scalability and generalizability of our RLRR method. As indicated in Tables~\ref{vitlhs_vtab}~(a) and (b), RLRR consistently outperforms other state-of-the-art adaptation methods, maintaining exceptional performance even when applied to these larger-scale backbones. Specifically, our method surpasses the latest state-of-the-art by 2.7\% on the ViT-L/16 and by 2.6\% on the ViT-H/14 backbones. These findings demonstrate the capability of RLRR to effectively scale to larger models, confirming its robustness for efficient adaptation across diverse Transformer-based architectures.

\noindent
\textbf{Experiments on hierarchical Vision Transformers. }
To further demonstrate the efficacy of RLRR, we apply it to the Swin Transformer~\cite{liu2021swin}, a Transformer-based architecture distinguished by its hierarchical structure. The Swin Transformer is organized into discrete stages, each with transformer blocks of consistent feature dimensions, though the dimensions vary across stages. Table~\ref{swinb_vtab} showcases that RLRR upholds competitive adaptation accuracy, even when adapted to this specialized Transformer architecture, thereby affirming its robustness to a range of visual adaptation tasks.

\begin{table}[!t]
	\centering
	\caption{Performance comparison on VTAB-1k using Swin Transformer pre-trained on ImageNet-21k as the backbone. "$(\cdot)$" indicates the number of tasks in the subgroup. Detailed results are presented in the Appendix.}
	\resizebox{\linewidth}{!}{
		\begin{tabular}{c|c|c|c|cc}
			\toprule
			\toprule
			\diagbox{\textbf{Methods}}{\textbf{Datasets}} & \textbf{Natural (7)} & \textbf{Specialized (4)} & \textbf{Structed (8)} & \textbf{Mean Total} & \textbf{Params.(M)} \\
			\midrule
			\midrule
			Full fine-tuning & 79.1  & 86.2  & \uline{59.7}  & 72.4  & 86.80  \\
			Linear probing & 73.5  & 80.8  & 33.5  & 58.2  & 0.05  \\
			\hline
			MLP-4~\cite{jia2022visual} & 70.6  & 80.7  & 31.2  & 57.7  & 4.04  \\
			Partial~\cite{jia2022visual} & 73.1  & 81.7  & 35.0  & 58.9  & 12.65  \\
			Bias~\cite{zaken2022bitfit}  & 74.2  & 80.1  & 42.4  & 62.1  & 0.25  \\
			VPT-Shallow~\cite{jia2022visual} & \uline{79.9}  & 82.5  & 37.8  & 62.9  & 0.05  \\
			VPT-Deep~\cite{jia2022visual} & 76.8  & 84.5  & 53.4  & 67.7  & 0.22  \\
			ARC~\cite{dong2023efficient}   & 79.0  & \uline{86.6}  & \textbf{59.9} & \uline{72.6}  & 0.27  \\
			\hline
			RLRR   & \textbf{81.3} & \textbf{86.7} & 59.0  & \textbf{73.0} & 0.41 \\
			\bottomrule
			\bottomrule
		\end{tabular}%
	}
	\label{swinb_vtab}%
\end{table}%

\begin{table}[!t]
	\centering
	\caption{Ablation study on the FGVC dataset to examine the impact of the various RLRR combinations.}
	\resizebox{\linewidth}{!}{
		\begin{tabular}{ccc|ccccc|cc}
			\toprule
			\toprule
			\multicolumn{2}{c}{\textbf{scaling}} & \multirow{2}[2]{*}{\textbf{residual}} & \multicolumn{5}{c|}{\textbf{FGVC Datasets}} & \multirow{2}[2]{*}{\textbf{Mean}} & \multicolumn{1}{c}{\multirow{2}[2]{*}{\textbf{Params. (M)}}} \\
			\textbf{left} & \textbf{right} &       & \multicolumn{1}{c}{\textbf{CUB}} & \textbf{NABirds} & \multicolumn{1}{c}{\textbf{Flowers}} & \multicolumn{1}{c}{\textbf{Dogs}} & \multicolumn{1}{c|}{\textbf{Cars}} &       &  \\
			\hline
			\checkmark & $\times$ & \multirow{3}[2]{*}{$\times$} & 86.9  & 84.2  & 99.5  & 91.0  & 84.3  & 89.2  & 0.39  \\
			$\times$ & \checkmark &       & 87.3  & 84.4  & 99.3  & 91.1  & 84.5  & 89.3  & 0.39  \\
			\checkmark & \checkmark &       & 86.6  & 83.9  & 99.3  & 91.0  & 83.5  & 88.9  & 0.47  \\
			\hline
			\checkmark & $\times$ & \multirow{3}[2]{*}{\checkmark} & 87.1  & 84.5  & 99.5  & 91.5  & 85.1  & 89.5  & 0.39  \\
			$\times$ & \checkmark &       & 87.9  & 84.5  & 99.4  & 91.3  & 85.4  & 89.7  & 0.39  \\
			\checkmark & \checkmark &       & 89.3  & 84.7  & 99.5  & 92.0  & 87.0  & 90.4  & 0.47  \\
			\bottomrule
			\bottomrule
		\end{tabular}
	}
	\label{tab_ablation}
	\vspace{-1.em}
\end{table}%

\subsection{Ablation Studies}
To gain deeper insights into the proposed method, we conduct comprehensive ablation studies on RLRR to elucidate its critical features and to carry out pertinent analyses. The ablation studies examining module deployment are performed using the CIFAR-100 dataset~\cite{krizhevsky2009learning}. Concurrently, we assess the impact of various components on FGVC dataset.

\begin{figure}[!t]
	\centering
	\includegraphics[width=1.0 \linewidth]{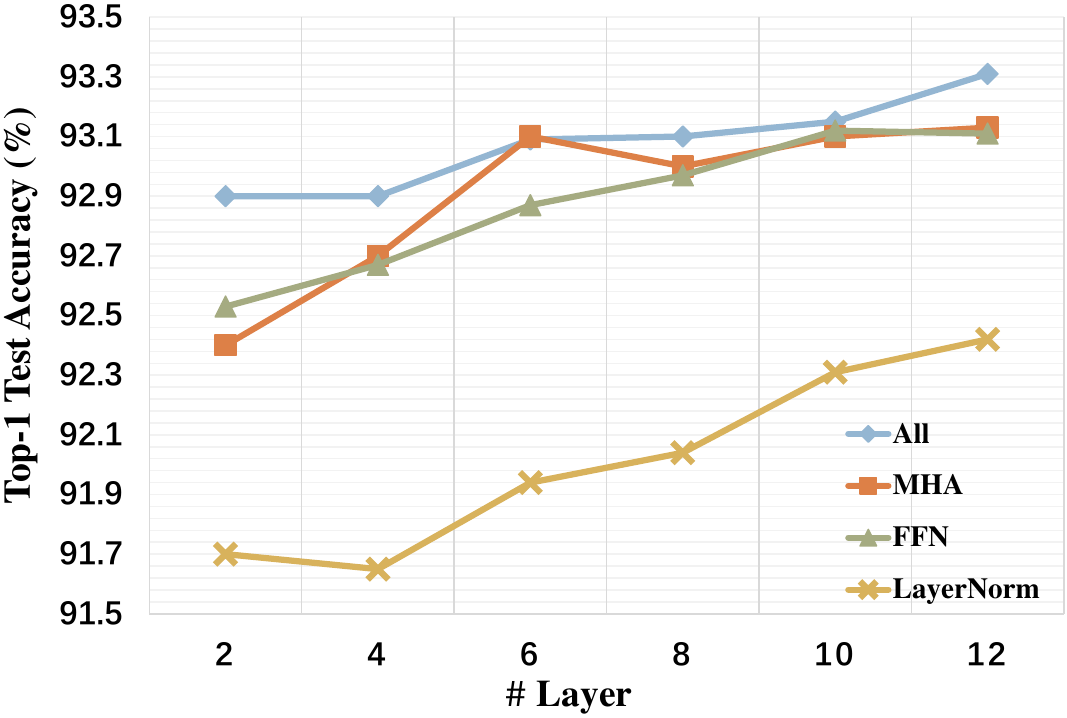}
	\caption{Ablation study using the VIT-B/16 backbone on the CIFAR-100 dataset to evaluate the impact of incorporating RLRR adaptation across different module and layer combinations.}
	\label{fig:ablation}
	\vspace{-1.em}
\end{figure}

\noindent
\textbf{Effect of RLRR Adaptation Insertion. }
To assess the impact of RLRR adaptation, we experiment with its insertion into different layers and Transformer modules, including MHA, FFN, and LayerNorm. Notably, for LayerNorm, as its weights are not stored in matrix form, we follow the same approach as SSF~\cite{lian2022scaling}. The specific results are illustrated in Fig.~\ref{fig:ablation}. We observe that as the number of deployed layers increases, the accuracy improves across all settings. Moreover, the configuration where the residual and rescaling design are applied to all modules, as we employed, consistently outperforms other configurations. Consequently, we choose to deploy the residual and rescaling design across all modules.

\noindent
\textbf{Effects of Different RLRR Combinations.}
To further illustrate the importance of the residual and rescaling design, we evaluate the ablation effects of the various components within our proposed method. The findings are delineated in Table~\ref{tab_ablation}. The results reveal that one-sided (\ie left or right) scaling tuning leads to better performance compared to dual-sided (\ie left and right) tuning in the absence of the residual term. This suggests that excessive rescaling of the original pre-trained parameter matrix will compromise the generalizability learned by pre-trained models, especially without additional constraints. Intriguingly, when the residual term is included, this trend reverses, which not only demonstrates that rescaling can introduce flexible perturbations but also emphasizes the importance of the residual term in maintaining the intrinsic representational capacity of the model.

%% file: sec/5_conclusion.tex
\section{Conclusions}
In this study, we addressed the challenge of PEFT for pre-trained Vision Transformers, with a focus on achieving a delicate balance between retaining the generalization capacity of the pre-trained model and adapting effectively to downstream tasks. Our approach involved viewing PEFT through a novel SVD perspective, offering a unified framework for understanding the working mechanisms of various PEFT strategies and their trade-offs.

To achieve a more favorable trade-off, we introduced a RLRR fine-tuning strategy. RLRR incorporates a residual term, providing enhanced adaptation flexibility while simultaneously preserving the representation capacity of the pre-trained model. Through extensive experiments on two downstream benchmark datasets, our RLRR method demonstrated highly competitive adaptation performance and exhibited other desirable properties. This work contributes valuable insights into the PEFT landscape and proposes an effective strategy for achieving a more nuanced balance between generalization and task-specific adaptation in pre-trained Vision Transformers.

%% file: sec/X_suppl.tex
\clearpage
\setcounter{page}{1}
\setcounter{section}{0}
\maketitlesupplementary

\setcounter{table}{0}
\begin{appendix}

\section{Detailed dataset statistic}
\label{sec:rationale}

We describe the details of visual adaptation classification tasks we used in Table~\ref{tab:dataset_fgvc}~(FGVC) and Table~\ref{tab:dataset_vtab}~(VTAB-1k), including the class number and the train/val/test sets. we employ the split following VPT~\cite{jia2022visual}.

\begin{table*}[!htbp] 
  \centering
  \caption{Dataset statistics for FGVC. ``*'' denotes the train/val split of datasets following the dataset setting in VPT~\cite{jia2022visual}.}
    \begin{tabular}{l|l|l|c|c|r}
    \toprule[2pt]
    \textbf{Dataset} & \textbf{Description} & \textbf{Classes} & \multicolumn{1}{l|}{\textbf{Train size}} & \multicolumn{1}{l|}{\textbf{Val size}} & \textbf{Test size} \\
    \hline
    CUB-200-2011~\cite{wah2011caltech} & Fine-grained bird species recognition & 200   & \multicolumn{1}{r|}{5,394*} & \multicolumn{1}{r|}{600*} & 5,794 \\
    NABirds~\cite{van2015building} & Fine-grained bird species recognition & 555    & \multicolumn{1}{r|}{21,536*} & \multicolumn{1}{r|}{2,393*} & 24,633 \\
    Oxford Flowers~\cite{nilsback2008automated} & Fine-grained flower species recognition & 102   & \multicolumn{1}{r|}{1,020} & \multicolumn{1}{r|}{1,020} & 6,149 \\
    Stanford Dogs~\cite{khosla2011novel} & Fine-grained dog species recognition & 120   & \multicolumn{1}{r|}{10,800*} & \multicolumn{1}{r|}{1,200*} & 8,580 \\
    Stanford Cars~\cite{gebru2017fine} & Fine-grained car classificatio & 196   & \multicolumn{1}{r|}{7,329*} & \multicolumn{1}{r|}{815*} & 8,041 \\
    \bottomrule[2pt]
    \end{tabular}%
  \label{tab:dataset_fgvc}
\end{table*}%

\begin{table*}[!htbp]
  \centering
  \caption{Dataset statistics for VTAB-1k~\cite{zhai2019large}.}
    \begin{tabular}{l|l|r|r|r|r}
    \toprule[2pt]
    \textbf{Dataset} & \textbf{Description} & \textbf{Classes} & \multicolumn{1}{r|}{\textbf{Train size}} & \multicolumn{1}{r|}{\textbf{Val size}} & \textbf{Test size} \\
    \hline
    CIFAR-100 
    & \multirow{7}[2]{*}{Natural} & 100   & \multirow{7}[2]{*}{800/1,000} & \multirow{7}[2]{*}{200} & 10,000 \\
    Caltech101 
    &       & 102   &       &       & 6,084 \\
    DTD 
    &       & 47    &       &       & 1,880 \\
    Flowers102 
    &       & 102   &       &       & 6,149 \\
    Pets 
    &       & 37    &       &       & 3,669 \\
    SVHN 
    &       & 10    &       &       & 26,032 \\
    Sun397 
    &       & 397   &       &       & 21,750 \\
    \hline
    Patch Camelyon 
    & \multirow{4}[1]{*}{Specialized} & 2     & \multirow{4}[1]{*}{800/1,000} & \multirow{4}[1]{*}{200} & 32,768 \\
    EuroSAT 
    &       & 10    &       &       & 5,400 \\
    Resisc45 
    &       & 45    &       &       & 6,300 \\
    Retinopathy 
    &       & 5     &       &       & 42,670 \\
    \hline
    Clevr/count 
    & \multirow{8}[1]{*}{Structured} & 8     & \multirow{8}[1]{*}{800/1,000} & \multirow{8}[1]{*}{200} & 15,000 \\
    Clevr/distance 
    &       & 6     &       &       & 15,000 \\
    DMLab 
    &       & 6     &       &       & 22,735 \\
    KITTI/distance 
    &       & 4     &       &       & 711 \\
    dSprites/location 
    &       & 16    &       &       & 73,728 \\
    dSprites/orientation 
    &       & 16    &       &       & 73,728 \\
    SmallNORB/azimuth 
    &       & 18    &       &       & 12,150 \\
    SmallNORB/elevation 
    &       & 9     &       &       & 12,150 \\
    \bottomrule[2pt]
    \end{tabular}%
  \label{tab:dataset_vtab}
\end{table*}%

\section{Detailed configuration}
Table~\ref{tab:optimizer_details} summarizes the detailed configurations we used for experiments. As mentioned in Section~\ref{sec:exp}, we utilize grid search to select hyper-parameters such as learning rate, weight decay, batch size, and dropout rate, using the validation set of each task. 
AugReg~\cite{steiner2021train} provides a robust initialization for the pre-training model with varying data augmentation and regularization. Despite the need for small initializations in many PEFT methods, RLRR maintains consistent performance under different initializations, as shown in Table~\ref{tab:init}. 

\begin{table*}[!htbp]
  \centering
  \caption{The implementation details of configurations such as optimizer and hyper-parameters. We select the best hyper-parameters for each download task via using grid search.}
    \begin{tabular}{c|c}
    \toprule[2pt]
    Optimizer & AdamW \\
    \hline
    Learning Rate & \{0.2, 0.1, 0.05, 0.01, 0.005, 0.001, 0.0001\} \\
    Weight Decay & \{0.05, 0.01, 0.005, 0.001, 0\} \\
    Dropout Rate & \{0, 0.1, 0.3, 0.5, 0.7\} \\
    Batch Size & \{256, 128, 32\} \\
    Learning Rate Schedule & Cosine Decay \\
    Training Epochs  & 100 \\
    Warmup Epochs & 10 \\
    \bottomrule[2pt]
    \end{tabular}%
  \label{tab:optimizer_details}%
\end{table*}%

\begin{table*}[!htbp]
  \renewcommand\arraystretch{1.3}
  \centering
  \caption{Comparison of the additional parameter size in both fine-tuning and inference stages with other lightweight adaptation methods.}
  \resizebox{2\columnwidth}{!}{
    \begin{tabular}{c|c|c|c|c|c|c|c}
    \toprule[2pt]
    \diagbox{\textbf{Stage}}{\textbf{Method}}
     & \textbf{Adapter~\cite{houlsby2019parameter}} & \textbf{VPT-Shallow~\cite{jia2022visual}} & \textbf{VPT-Deep~\cite{jia2022visual}} & \textbf{LoRA~\cite{hu2021lora}}  & \textbf{SSF~\cite{lian2022scaling}}   & \textbf{ARC~\cite{dong2023efficient}} & \textbf{RLRR} \\
    \hline
    Fine-Tuning &  \(2\cdot D\cdot D^\prime \cdot L\) & \(m\cdot D\)    & \({m\cdot D\cdot L}\)     & \({2\cdot w\cdot D\cdot D^\prime \cdot L}\)      & \(2\cdot o\cdot D^* \cdot L\)     & \(2\cdot (D\cdot D^\prime +(D^\prime +D)\cdot L)\)   & \(3\cdot o\cdot D^* \cdot L\) \\
    Inference &  \(2\cdot D\cdot D^\prime \cdot L\) & \(m\cdot D\)   & \({m\cdot D\cdot L}\)     & 0     & 0     & 0 & 0\\
    \bottomrule[2pt]
    \end{tabular}%
    }
  \label{tab:params_analysis_comprision}
\end{table*}

\begin{table*}[!htbp]
  \centering
  \caption{This table is extended from Table~\ref{vitlhs_vtab} in Section~\ref{sec:exp} and describes the detailed experimental results of the performance comparison on VTAB-1k using ViT-Large pre-trained on ImageNet-21k as the backbone.}
  \resizebox{\linewidth}{!}{
    \begin{tabular}{c|ccccccc|c|cccc|c|cccccccc|c|cc}
    \toprule
    \toprule
    \multirow{2}[2]{*}{\diagbox{\textbf{Methods}}{\textbf{Datasets}}} & \multicolumn{8}{c|}{\textbf{Natural}}                         & \multicolumn{5}{c|}{\textbf{Specialized}} & \multicolumn{9}{c|}{\textbf{Structed}}                                &       &  \\
          & \rotatebox{90}{\textbf{CIFAR-100}} & \rotatebox{90}{\textbf{Caltech101}} & \rotatebox{90}{\textbf{DTD}} & \rotatebox{90}{\textbf{Flowers102}} & \rotatebox{90}{\textbf{Pets}} & \rotatebox{90}{\textbf{SVNH}} & \multicolumn{1}{c}{\rotatebox{90}{\textbf{Sun397}}} & \rotatebox{90}{\textbf{Mean}} & \rotatebox{90}{\textbf{Camelyon}} & \rotatebox{90}{\textbf{EuroSAT}} & \rotatebox{90}{\textbf{Resisc45}} & \multicolumn{1}{c}{\rotatebox{90}{\textbf{Retinopathy}}} & \rotatebox{90}{\textbf{Mean}} & \rotatebox{90}{\textbf{Clevr-Count}} & \rotatebox{90}{\textbf{Clevr-Dist}} & \rotatebox{90}{\textbf{DMLab}} & \rotatebox{90}{\textbf{KITTI-Dist}} & \rotatebox{90}{\textbf{dSpr-Loc}} & \rotatebox{90}{\textbf{dSpr-Ori}} & \rotatebox{90}{\textbf{sNORB-Azim}} & \multicolumn{1}{c}{\rotatebox{90}{\textbf{sNORB-Ele}}} & \rotatebox{90}{\textbf{Mean}} & \rotatebox{90}{\textbf{Mean Total}} & \rotatebox{90}{\textbf{Params.(M)}} \\
    \midrule
    Full fine-tuning & 68.6  & 84.3  & 58.6  & 96.3  & 86.5  & 87.5  & 41.4  & 74.7  & 82.6  & \uline{95.9} & 82.4  & 74.2  & 83.8  & 55.4  & 55.0  & 42.2  & 74.2  & 56.8  & 43.0  & 28.5  & 29.7  & 48.1  & 65.4  & 303.4 \\
    Linear probing & 72.2  & 86.4  & 63.6  & 97.4  & 85.8  & 38.1  & 52.5  & 70.9  & 76.9  & 87.3  & 66.6  & 45.4  & 69.1  & 28.2  & 28.0  & 34.7  & 54.0  & 10.6  & 14.2  & 14.6  & 21.9  & 25.8  & 51.5  & 0.05 \\
    \midrule
    Adapter~\cite{houlsby2019parameter} & 75.3  & 84.2  & 54.5  & 97.4  & 84.3  & 31.3  & 52.9  & 68.6  & 75.8  & 85.1  & 63.4  & 69.5  & 73.5  & 35.4  & 34.1  & 30.8  & 47.1  & 30.4  & 23.4  & 10.8  & 19.8  & 29.0  & 52.9  & 2.38 \\
    Bias~\cite{zaken2022bitfit} & 71.0  & 82.4  & 51.3  & 96.3  & 83.2  & 59.5  & 49.9  & 70.5  & 72.9  & 87.9  & 63.1  & 71.3  & 73.8  & 51.2  & 50.7  & 33.5  & 54.8  & 65.9  & 37.3  & 13.7  & 22.2  & 41.2  & 58.9  & 0.32 \\
    VPT-Shallow~\cite{jia2022visual} & \uline{80.6} & 88.2  & 67.1  & 98.0  & 85.9  & 78.4  & 53.0  & 78.7  & 79.7  & 93.5  & 73.4  & 73.1  & 79.9  & 41.5  & 52.5  & 32.3  & 64.2  & 48.3  & 35.3  & 21.6  & 28.8  & 40.6  & 62.9  & 0.15 \\
    VPT-Deep~\cite{jia2022visual} & \textbf{84.1} & 88.9  & 70.8  & 98.8  & 90.0  & 89.0  & 55.9  & \uline{82.5} & 82.5  & \textbf{96.6} & 82.6  & 73.9  & 83.9  & 63.7  & \uline{60.7}  & 46.1  & 75.7  & \uline{83.7} & 47.4  & 18.9  & \uline{36.9} & 54.1  & 70.8  & 0.49 \\
    LoRA~\cite{hu2021lora} & 75.8  & 89.8 & \uline{73.6} & 99.1 & 90.8 & 83.2  & \uline{57.5} & 81.4  & \uline{86.0} & 95.0  & 83.4  & 75.5  & 85.0  & 78.1  & 60.5 & 46.7 & \textbf{81.6} & 76.7  & 51.3  & 28.0  & 35.4  & 57.3 & 72.0    & 0.74 \\
    ARC~\cite{dong2023efficient} & 76.2  & 89.6  & 73.4  & 99.1 & 90.3  & \textbf{90.9} & 56.5  & 82.3  & 85.0  & 95.7  & \uline{85.9} & \textbf{75.8} & \uline{85.6} & 78.6 & \textbf{62.1} & 46.7 & 76.7  & 75.9  & 53.0 & 30.2  & 35.2  & 57.3 & 72.5 & 0.18 \\
    SSF~\cite{lian2022scaling} & 73.5  & \uline{91.3}  & 70.0  & \uline{99.3} & \uline{91.3}  & \uline{90.6} & \uline{57.5}  & 81.9  & 85.9  & 94.9  & 85.5 & 74.4 & 85.2 & \uline{80.6} & 60.0 & \uline{53.3} & 80.0  & 77.6  & \uline{54.0} & \textbf{31.8} & 35.0  & \uline{59.0} & \uline{73.0} & 0.60 \\
    \midrule
    RLRR  & 79.3  & \textbf{92.0} & \textbf{74.6} & \textbf{99.5} & \textbf{92.1} & 89.6 & \textbf{60.1} & \textbf{83.9} & \textbf{87.3} & 95.3  & \textbf{87.3} & \uline{75.7} & \textbf{86.4} & \textbf{82.7} & \textbf{62.1} & \textbf{54.6} & \uline{80.6} & \textbf{87.1} & \textbf{54.7} & \uline{31.3} & \textbf{41.9} & \textbf{61.9} & \textbf{75.2} & 0.82 \\
    \bottomrule
    \bottomrule
    \end{tabular}%
    }
  \label{tab:Ldetail}
\end{table*}

\begin{table*}[!htbp]
  \centering
  \caption{This table is extended from Table~\ref{vitlhs_vtab} in Section~\ref{sec:exp} and describes the detailed experimental results of the performance comparison on VTAB-1k using ViT-Huge pre-trained on ImageNet-21k as the backbone.}
  \resizebox{\linewidth}{!}{
        \begin{tabular}{c|ccccccc|c|cccc|c|cccccccc|c|cc}
    \toprule
    \toprule
    \multirow{2}[2]{*}{\diagbox{\textbf{Methods}}{\textbf{Datasets}}} & \multicolumn{8}{c|}{\textbf{Natural}}    & \multicolumn{5}{c|}{\textbf{Specialized}} & \multicolumn{9}{c|}{\textbf{Structed}}                                &       &  \\
          & \rotatebox{90}{\textbf{CIFAR-100}} & \rotatebox{90}{\textbf{Caltech101}} & \rotatebox{90}{\textbf{DTD}} & \rotatebox{90}{\textbf{Flowers102}} & \rotatebox{90}{\textbf{Pets}} & \rotatebox{90}{\textbf{SVNH}} & \multicolumn{1}{c}{\rotatebox{90}{\textbf{Sun397}}} & \rotatebox{90}{\textbf{Mean}} & \rotatebox{90}{\textbf{Camelyon}} & \rotatebox{90}{\textbf{EuroSAT}} & \rotatebox{90}{\textbf{Resisc45}} & \multicolumn{1}{c}{\rotatebox{90}{\textbf{Retinopathy}}} & \rotatebox{90}{\textbf{Mean}} & \rotatebox{90}{\textbf{Clevr-Count}} & \rotatebox{90}{\textbf{Clevr-Dist}} & \rotatebox{90}{\textbf{DMLab}} & \rotatebox{90}{\textbf{KITTI-Dist}} & \rotatebox{90}{\textbf{dSpr-Loc}} & \rotatebox{90}{\textbf{dSpr-Ori}} & \rotatebox{90}{\textbf{sNORB-Azim}} & \multicolumn{1}{c}{\rotatebox{90}{\textbf{sNORB-Ele}}} & \rotatebox{90}{\textbf{Mean}} & \rotatebox{90}{\textbf{Mean Total}} & \rotatebox{90}{\textbf{Params.(M)}} \\
    \midrule
    Full fine-tuning & 58.7  & 86.5  & 55.0  & 96.5  & 79.7  & 87.5  & 32.5  & 70.9  & 83.1  & \uline{95.5}  & 81.9  & 73.8  & 83.6  & 47.6  & 53.9  & 37.8  & 69.9  & 53.8  & 48.6  & 30.2  & 25.8  & 46.0  & 63.1  & 630.90  \\
    Linear probing & 64.3  & 83.6  & 65.2  & 96.2  & 83.5  & 39.8  & 43.0  & 67.9  & 78.0  & 90.5  & 73.9  & 73.4  & 79.0  & 25.6  & 24.5  & 34.8  & 59.0  & 9.5   & 15.6  & 17.4  & 22.8  & 26.1  & 52.7  & 0.06  \\
    \midrule
    Adapter~\cite{houlsby2019parameter} & 69.4  & 84.4  & 62.7  & 97.2  & 84.2  & 33.6  & 45.3  & 68.1  & 77.3  & 86.6  & 70.8  & 71.1  & 76.4  & 28.6  & 27.5  & 29.2  & 55.2  & 10.0  & 15.2  & 11.9  & 18.6  & 24.5  & 51.5  & 5.78  \\
    Bias~\cite{zaken2022bitfit} & 65.7  & 84.3  & 59.9  & 96.6  & 80.6  & 60.1  & 44.9  & 70.3  & 79.7  & 92.8  & 71.5  & 71.6  & 78.9  & 52.3  & 50.4  & 31.2  & 57.7  & 65.9  & 39.7  & 16.7  & 20.2  & 41.7  & 60.1  & 0.52  \\
    VPT-Shallow~\cite{jia2022visual} & \uline{70.6} & 84.7  & 64.8  & 96.4  & 85.1  & 75.6  & 46.2  & 74.8  & 79.9  & 93.7  & 77.7  & 73.6  & 81.2  & 40.3  & 60.9  & 34.9  & 63.3  & 61.3  & 38.9  & 19.8  & 24.9  & 43.0  & 62.8  & 0.18  \\
    VPT-Deep~\cite{jia2022visual} & \textbf{76.9} & 87.2  & 66.8  & 97.5  & 84.8  & 85.5  & 46.5  & 77.9  & 81.6  & \textbf{96.3} & 82.5  & 72.8  & 83.3  & 50.4  & 61.2  & 43.9  & \uline{76.6}  & 79.5 & 50.1  & 24.7  & 31.5  & 52.2  & 68.2  & 0.96  \\
    LoRA~\cite{hu2021lora} & 63.0  & 89.4  & 68.1  & 98.0  & 87.0  & 85.2  & 48.7  & 77.1  & 82.2  & 94.3  & 83.1  & \uline{74.2} & 83.5  & 68.6  & \uline{65.0}  & 44.8  & 76.4 & 70.8  & 48.8  & 30.4  & \uline{38.3} & 55.4 & 69.3  & 1.21  \\
    ARC~\cite{dong2023efficient} & 67.6  & \uline{90.2} & \uline{69.5} & \uline{98.4} & \uline{87.9} & \textbf{90.8} & 49.6 & \uline{79.1} & 84.5 & 94.9  & \textbf{85.1} & \textbf{74.6} & \uline{84.8} & \textbf{75.2} & \textbf{66.7} & 46.2 & 76.4 & 44.2  & 51.1 & \textbf{32.2} & 37.7  & 53.7  & 69.6 & 0.22  \\
    SSF~\cite{lian2022scaling} & 66.6  & \textbf{91.2} & 69.0 & \uline{98.4} & \textbf{88.1} & \uline{88.9} & \uline{50.7} & 79.0 & \uline{85.0} & 94.1  & 79.3 & 73.9 & 83.1 & \uline{73.9} & 61.2 & \uline{47.9} & 76.2 & \uline{82.8}  & \uline{51.9} & 25.5 & 33.7  & \uline{56.6}  & \uline{70.4} & 0.97  \\
    \midrule
    RLRR  & 70.3  & 89.8 & \textbf{69.7} & \textbf{98.6} & 87.8 & 88.5 & \textbf{51.3} & \textbf{79.4} & \textbf{86.0} & 95.0 & \uline{84.9} & \textbf{74.6} & \textbf{85.1} & 73.8 & 60.1 & \textbf{49.6} & \textbf{78.6} & \textbf{83.6} & \textbf{52.4} & \uline{32.0} & \textbf{41.8} & \textbf{59.0} & \textbf{72.0} & 1.33 \\
    \bottomrule
    \bottomrule
    \end{tabular}%
    }
  \label{tab:Hdetail}
\end{table*}

\begin{table*}[!htbp]
  \centering
  \caption{This table is extended from Table~\ref{swinb_vtab} in Section~\ref{sec:exp} and describes the detailed experimental results of the performance comparison on VTAB-1k using Swin-Base pre-trained on ImageNet-21k as the backbone.}
  \resizebox{\linewidth}{!}{
    \begin{tabular}{c|ccccccc|c|cccc|c|cccccccc|c|cc}
    \toprule
    \toprule
    \multirow{2}[2]{*}{\diagbox{\textbf{Methods}}{\textbf{Datasets}}} & \multicolumn{8}{c|}{\textbf{Natural}}                         & \multicolumn{5}{c|}{\textbf{Specialized}} & \multicolumn{9}{c|}{\textbf{Structed}}                                &       &  \\
          & \rotatebox{90}{\textbf{CIFAR-100}} & \rotatebox{90}{\textbf{Caltech101}} & \rotatebox{90}{\textbf{DTD}} & \rotatebox{90}{\textbf{Flowers102}} & \rotatebox{90}{\textbf{Pets}} & \rotatebox{90}{\textbf{SVNH}} & \multicolumn{1}{c}{\rotatebox{90}{\textbf{Sun397}}} & \rotatebox{90}{\textbf{Mean}} & \rotatebox{90}{\textbf{Camelyon}} & \rotatebox{90}{\textbf{EuroSAT}} & \rotatebox{90}{\textbf{Resisc45}} & \multicolumn{1}{c}{\rotatebox{90}{\textbf{Retinopathy}}} & \rotatebox{90}{\textbf{Mean}} & \rotatebox{90}{\textbf{Clevr-Count}} & \rotatebox{90}{\textbf{Clevr-Dist}} & \rotatebox{90}{\textbf{DMLab}} & \rotatebox{90}{\textbf{KITTI-Dist}} & \rotatebox{90}{\textbf{dSpr-Loc}} & \rotatebox{90}{\textbf{dSpr-Ori}} & \rotatebox{90}{\textbf{sNORB-Azim}} & \multicolumn{1}{c}{\rotatebox{90}{\textbf{sNORB-Ele}}} & \rotatebox{90}{\textbf{Mean}} & \rotatebox{90}{\textbf{Mean Total}} & \rotatebox{90}{\textbf{Params.(M)}} \\
    \midrule
    Full fine-tuning & 72.2  & 88.0  & 71.4  & 98.3  & 89.5  & 89.4  & 45.1  & 79.1  & 86.6  & \textbf{96.9} & \textbf{87.7} & 73.6  & 86.2  & \textbf{75.7} & \textbf{59.8} & \uline{54.6} & 78.6  & 79.4  & \uline{53.6} & \textbf{34.6} & \textbf{40.9} & \uline{59.7} & 72.4  & 86.9 \\
    Linear probing & 61.4  & 90.2  & 74.8  & 95.5  & 90.2  & 46.9  & \textbf{55.8} & 73.5  & 81.5  & 90.1  & 82.1  & 69.4  & 80.8  & 39.1  & 35.9  & 40.1  & 65.0  & 20.3  & 26.0  & 14.3  & 27.6  & 33.5  & 58.2  & 0.05 \\
    \midrule
    MLP-4~\cite{jia2022visual} & 54.9  & 87.4  & 71.4  & 99.5  & 89.1  & 39.7  & 52.5  & 70.6  & 80.5  & 90.9  & 76.8  & 74.4  & 80.7  & 60.9  & 38.8  & 40.2  & 66.5  & 9.4   & 21.1  & 14.5  & 28.8  & 31.2  & 57.7  & 4.04 \\
    Partial~\cite{jia2022visual} & 60.3  & 88.9  & 72.6  & 98.7  & 89.3  & 50.5  & 51.5  & 73.1  & 82.8  & 91.7  & 80.1  & 72.3  & 81.7  & 34.3  & 35.5  & 43.2  & 77.1  & 15.8  & 26.2  & 19.1  & 28.4  & 35.0  & 58.9  & 12.65 \\
    Bias~\cite{zaken2022bitfit}  & 73.1  & 86.8  & 65.7  & 97.7  & 87.5  & 56.4  & 52.3  & 74.2  & 80.4  & 91.6  & 76.1  & 72.5  & 80.1  & 47.3  & 48.5  & 34.7  & 66.3  & 57.6  & 36.2  & 17.2  & 31.6  & 42.4  & 62.1  & 0.25 \\
    VPT-Shallow~\cite{jia2022visual} & \uline{78.0} & \textbf{91.3} & \uline{77.2} & \uline{99.4} & 90.4  & 68.4  & 54.3  & \uline{79.9}  & 80.1  & 93.9  & 83.0  & 72.7  & 82.5  & 40.8  & 43.9  & 34.1  & 63.2  & 28.4  & 44.5  & 21.5  & 26.3  & 37.8  & 62.9  & 0.05 \\
    VPT-Deep~\cite{jia2022visual} & \textbf{79.6} & \uline{90.8} & \textbf{78.0} & \textbf{99.5} & \uline{91.4} & 46.5  & 51.7  & 76.8  & 84.9  & \uline{96.2} & 85.0  & 72.0  & 84.5  & 67.6  & \uline{59.4} & 50.1  & 74.1  & 74.4  & 50.6  & 25.7  & 25.7  & 53.4  & 67.7  & 0.22 \\
    ARC~\cite{dong2023efficient}   & 62.5  & 90.0  & 71.9  & 99.2  & 87.8  & \uline{90.7} & 51.1  & 79.0 & \textbf{89.1} & 95.8  & 84.5  & \textbf{77.0} & \uline{86.6} & \uline{75.4} & 57.4  & 53.4  & \uline{83.1} & \textbf{91.7} & \textbf{55.2} & \uline{31.6} & 31.8  & \textbf{59.9} & \uline{72.6} & 0.27 \\
    \midrule
    RLRR  & 66.1  & 90.6  & 75.5  & 99.3  & \textbf{92.1} & \textbf{90.9} & \uline{54.7} & \textbf{81.3} & \uline{87.1} & 95.9  & \uline{87.1} & \uline{76.5} & \textbf{86.7} & 66.0  & 57.8  & \textbf{55.3} & \textbf{84.1} & \uline{91.1} & \textbf{55.2} & 28.6  & \uline{34.0} & 59.0  & \textbf{73.0} & 0.41 \\
    \bottomrule
    \bottomrule
    \end{tabular}%
    }
  \label{tab:Swindetail}
\end{table*}

\begin{table*}[!h]
  \centering
  \caption{Performance comparison on VTAB-1k using MAE self-supervised pre-trained ViT-Base as backbone.}
  \resizebox{\linewidth}{!}{
    \begin{tabular}{c|ccccccc|c|cccc|c|cccccccc|c|cc}
    \toprule
    \toprule
    \multirow{2}[2]{*}{\diagbox{\textbf{Methods}}{\textbf{\rotatebox{0}{Datasets}}}} & \multicolumn{8}{c|}{\textbf{Natural}}                         & \multicolumn{5}{c|}{\textbf{Specialized}} & \multicolumn{9}{c|}{\textbf{Structed}}                                &       &  \\
          & \rotatebox{90}{\textbf{CIFAR-100}} & \rotatebox{90}{\textbf{Caltech101}} & \rotatebox{90}{\textbf{DTD}} & \rotatebox{90}{\textbf{Flowers102}} & \rotatebox{90}{\textbf{Pets}} & \rotatebox{90}{\textbf{SVNH}} & \multicolumn{1}{c}{\rotatebox{90}{\textbf{Sun397}}} & \rotatebox{90}{\textbf{Mean}} & \rotatebox{90}{\textbf{Camelyon}} & \rotatebox{90}{\textbf{EuroSAT}} & \rotatebox{90}{\textbf{Resisc45}} & \multicolumn{1}{c}{\rotatebox{90}{\textbf{Retinopathy}}} & \rotatebox{90}{\textbf{Mean}} & \rotatebox{90}{\textbf{Clevr-Count}} & \rotatebox{90}{\textbf{Clevr-Dist}} & \rotatebox{90}{\textbf{DMLab}} & \rotatebox{90}{\textbf{KITTI-Dist}} & \rotatebox{90}{\textbf{dSpr-Loc}} & \rotatebox{90}{\textbf{dSpr-Ori}} & \rotatebox{90}{\textbf{sNORB-Azim}} & \multicolumn{1}{c}{\rotatebox{90}{\textbf{sNORB-Ele}}} & \rotatebox{90}{\textbf{Mean}} & \rotatebox{90}{\textbf{Mean Total}} & \rotatebox{90}{\textbf{Params.(M)}} \\
    \midrule
    \midrule
    Full fine tuning & 24.6  & 84.2  & 56.9  & 72.7  & 74.4  & 86.6  & 15.8  & 59.3  & 81.8  & \textbf{94.0} & 72.3  & 70.6  & 79.7  & 67.0  & 59.8  & 45.2  & 75.3  & 72.5  & 47.5  & 30.2  & 33.0  & 53.8  & 61.3  & 85.80  \\
    Linear & 8.7   & 41.5  & 20.6  & 19.2  & 11.3  & 22.3  & 8.6   & 18.9  & 76.5  & 68.6  & 16.6  & 53.2  & 53.7  & 33.6  & 32.5  & 23.0  & 51.1  & 13.0  & 9.9   & 8.5   & 17.9  & 23.7  & 28.2  & 0.04  \\
    \hline
    Bias~\cite{zaken2022bitfit}  & 22.4  & 82.6  & 49.7  & 66.2  & 67.7  & 69.0  & 24.3  & 54.6  & 78.7  & 91.4  & 60.0  & 72.6  & 75.7  & 65.9  & 51.0  & 35.0  & 69.1  & 70.8  & 37.6  & 21.5  & 30.7  & 47.7  & 56.1  & 0.14  \\
    Adapter~\cite{houlsby2019parameter} & \textbf{35.1} & 85.0  & 56.5  & 66.6  & 71.3  & 45.0  & \uline{24.8} & 54.9  & 76.9  & 87.1  & 63.5  & 73.3  & 75.2  & 43.8  & 49.5  & 31.2  & 61.7  & 59.3  & 23.3  & 13.6  & 29.6  & 39.0  & 52.5  & 0.76  \\
    VPT-Shallow~\cite{jia2022visual} & 21.9  & 76.2  & 54.7  & 58.0  & 41.3  & 16.1  & 15.1  & 40.0  & 74.0  & 69.5  & 58.9  & 72.7  & 68.8  & 40.3  & 44.7  & 27.9  & 60.5  & 11.8  & 11.0  & 12.4  & 16.3  & 28.1  & 41.2  & 0.04  \\
    VPT-Deep~\cite{jia2022visual} & 8.2   & 55.2  & 58.0  & 39.3  & 45.2  & 19.4  & 21.9  & 35.3  & 77.9  & 91.0  & 45.4  & 73.6  & 72.0  & 39.0  & 40.9  & 30.6  & 53.9  & 21.0  & 12.1  & 11.0  & 14.9  & 27.9 & 39.9  & 0.06  \\
    LoRA~\cite{hu2021lora}  & 31.8  & 88.4  & 59.9  & 81.7  & \uline{85.3} & \uline{90.3}  & 23.7  & 65.9  & \uline{84.2}  & 92.5  & 76.2  & \textbf{75.4}  & 82.1  & \textbf{85.9} & 64.1  & 49.4  & \uline{82.8}  & \textbf{83.9}  & 51.8  & 34.6 & \uline{41.3}  & \uline{61.7}  & 67.5  & 0.30  \\
    ARC~\cite{dong2023efficient}   & 31.3  & \textbf{89.3} & \uline{61.2}  & \uline{85.9} & 83.1  & \textbf{91.6} & 24.4  & \uline{66.7}  & \textbf{86.0} & \textbf{94.0} & \uline{80.4}  & 74.8  & \uline{83.8}  & \uline{85.8}  & \uline{64.6} & \uline{50.5}  & \uline{82.8}  & \uline{82.8}  & \uline{53.5}  & \uline{36.3} & 39.7  & \textbf{62.0}  & \uline{68.3}  & 0.13  \\
    \hline
    RLRR & \uline{33.6} & \uline{88.9} & \textbf{62.2} & \textbf{87.3} & \textbf{86.7} & 89.1 & \textbf{25.7} & \textbf{67.6} & \textbf{86.0} & \uline{93.4} & \textbf{81.3} & \uline{75.1} & \textbf{84.0} & 77.0 & \textbf{65.5} & \textbf{53.4} & \textbf{84.7} & 78.5 & \textbf{54.5} & \textbf{37.2} & \textbf{43.1} & \uline{61.7} & \textbf{68.6} & 0.33
    \\
    \bottomrule
    \bottomrule
    \end{tabular}
    }
  \label{tab:Maedetail}
\end{table*}

\begin{table*}[!t]
  \centering
  \caption{Performance comparison on VTAB-1k using Moco V3 self-supervised pre-trained ViT-Base as backbone.}
  \resizebox{\linewidth}{!}{
    \begin{tabular}{c|ccccccc|c|cccc|c|cccccccc|c|cc}
    \toprule
    \toprule
    \multirow{2}[2]{*}{\diagbox{\textbf{Methods}}{\textbf{\rotatebox{0}{Datasets}}}} & \multicolumn{8}{c|}{\textbf{Natural}}                         & \multicolumn{5}{c|}{\textbf{Specialized}} & \multicolumn{9}{c|}{\textbf{Structed}}                                &       &  \\
          & \rotatebox{90}{\textbf{CIFAR-100}} & \rotatebox{90}{\textbf{Caltech101}} & \rotatebox{90}{\textbf{DTD}} & \rotatebox{90}{\textbf{Flowers102}} & \rotatebox{90}{\textbf{Pets}} & \rotatebox{90}{\textbf{SVNH}} & \multicolumn{1}{c}{\rotatebox{90}{\textbf{Sun397}}} & \rotatebox{90}{\textbf{Mean}} & \rotatebox{90}{\textbf{Camelyon}} & \rotatebox{90}{\textbf{EuroSAT}} & \rotatebox{90}{\textbf{Resisc45}} & \multicolumn{1}{c}{\rotatebox{90}{\textbf{Retinopathy}}} & \rotatebox{90}{\textbf{Mean}} & \rotatebox{90}{\textbf{Clevr-Count}} & \rotatebox{90}{\textbf{Clevr-Dist}} & \rotatebox{90}{\textbf{DMLab}} & \rotatebox{90}{\textbf{KITTI-Dist}} & \rotatebox{90}{\textbf{dSpr-Loc}} & \rotatebox{90}{\textbf{dSpr-Ori}} & \rotatebox{90}{\textbf{sNORB-Azim}} & \multicolumn{1}{c}{\rotatebox{90}{\textbf{sNORB-Ele}}} & \rotatebox{90}{\textbf{Mean}} & \rotatebox{90}{\textbf{Mean Total}} & \rotatebox{90}{\textbf{Params.(M)}} \\
    \midrule
    \midrule
    Full fine tuning & 57.6  & 91.0  & 64.6  & 91.5  & 79.9  & 89.8  & 29.1  & 72.0  & 85.1  & \textbf{96.4} & 83.1  & 74.3  & 84.7  & 55.1  & 56.9  & 44.7  & 77.9  & 63.8  & 49.0  & 31.5  & 36.9  & 52.0  & 66.2  & 85.69  \\
    Linear & 62.9  & 85.1  & 68.8  & 87.0  & 85.8  & 41.8  & \uline{40.9} & 67.5  & 80.3  & 93.6  & 77.9  & 72.6  & 81.1  & 42.3  & 34.8  & 36.4  & 59.2  & 10.1  & 22.7  & 12.6  & 24.7  & 30.3  & 54.7  & 0.04  \\
    \hline
    Bias~\cite{zaken2022bitfit}  & 65.5  & 89.2  & 62.9  & 88.9  & 80.5  & 82.7  & 40.5  & 72.9  & 80.9  & 95.2  & 77.7  & 70.8  & 81.1  & 71.4  & 59.4  & 39.8  & 77.4  & 70.2  & 49.0  & 17.5  & 42.8  & 53.4  & 66.4  & 0.14  \\
    Adapter~\cite{houlsby2019parameter} & \textbf{73.0} & 88.2  & \uline{69.3}  & 90.7  & 87.4  & 69.9  & \uline{40.9} & 74.2  & 82.4  & 93.4  & 80.5  & 74.3  & 82.7  & 55.6  & 56.1  & 39.1  & 73.9  & 60.5  & 40.2  & 19.0  & 37.1  & 47.7  & 64.8  & 0.98  \\
    VPT-Shallow~\cite{jia2022visual} & 68.3  & 86.8  & \textbf{69.7} & 90.0  & 59.7  & 56.9  & 39.9  & 67.3  & 81.7  & 94.7  & 78.9  & 73.8  & 82.3  & 34.3  & 56.8  & 40.6  & 49.1  & 40.4  & 31.8  & 13.1  & 34.4  & 37.6  & 57.9  & 0.05  \\
    VPT-Deep~\cite{jia2022visual} & \uline{70.1}  & 88.3  & 65.9  & 88.4  & 85.6  & 57.8  & 35.7  & 70.3  & 83.1  & 93.9  & 81.2  & 74.0  & 83.0  & 48.5  & 55.8  & 37.2  & 64.6  & 52.3  & 26.5  & 19.4  & 34.8  & 42.4  & 61.2  & 0.05  \\
    LoRA~\cite{hu2021lora}  & 58.8  & 90.8  & 66.0  & \uline{91.8}  & 88.1  & 87.6  & 40.6  & 74.8  & 86.4  & 95.3  & 83.4  & \uline{75.5}  & \uline{85.1}  & \textbf{83.0}  & \textbf{64.6}  & \uline{51.3}  & \textbf{81.9}  & \uline{83.2}  & 47.5  & 32.4  & \uline{47.3}  & 61.4  & 71.3  & 0.30  \\
    ARC~\cite{dong2023efficient}   & 60.0  & \uline{91.3} & 67.9  & \textbf{92.8}  & \uline{89.3} & \uline{91.4} & \uline{40.9} & \uline{76.2} & \uline{87.5} & 95.6  & \textbf{86.1} & \textbf{75.6} & \textbf{86.2} & \textbf{83.0}  & \uline{64.2}  & 50.2  & 80.6  & \textbf{85.0}  & \textbf{53.0}  & \uline{34.6} & \textbf{47.4} & \textbf{62.3}  & \uline{72.4}  & 0.13  \\
    \hline
    RLRR & 61.8 & \textbf{91.7} & 68.6 & 91.6 & \textbf{89.5} & \textbf{91.5} & \textbf{41.7} & \textbf{76.6} & \textbf{87.9} & \uline{96.0} & \uline{85.4} & 75.4 & \textbf{86.2} & \uline{79.3} & \textbf{64.6} & \textbf{51.5} & \uline{81.4} & 77.5 & \uline{50.4} & \textbf{35.6} & 45.9 & \uline{62.1} & \textbf{73.1} & 0.33
    \\
    \bottomrule
    \bottomrule
    \end{tabular}
    }
  \label{tab:Mocodetail}
\end{table*}

\begin{table}[htbp]
  \centering
  \caption{The impacts of initialization.}
  \resizebox{\linewidth}{!}{
    \begin{tabular}{c|ccc}
    \toprule
    Initialization & Natural (7) & Specialized (4) & Structed (8) \\
    \midrule
    normal & \textbf{82.9 } & 85.4  & 61.4  \\
    zero  & 82.4  & 85.1  & 60.8  \\
    constant & 81.6  & 84.2  & 60.9  \\
    uniform & 82.5  & 85.6  & 61.4  \\
    RLRR  & 82.7  & \textbf{85.8} & \textbf{61.8} \\
    \bottomrule
    \end{tabular}%
    }
  \label{tab:init}%
\end{table}%

\section{Parameter size analysis}
To showcase the parameter-efficiency of our RLRR method, we compare its parameter size with other popular lightweight adaptation methods (Table~\ref{tab:params_analysis_comprision}), including Adapter~\cite{houlsby2019parameter}, VPT~\cite{jia2022visual}, LoRA~\cite{hu2021lora}, SSF~\cite{lian2022scaling} and ARC~\cite{dong2023efficient}. Adapter~\cite{houlsby2019parameter} uses two linear projections to construct a bottleneck structure for each layer, resulting in the introduction of \(2\cdot D\cdot D^\prime\cdot L\) learnable parameters, where \(D^\prime\) denotes the size of hidden dimension and \(L\) denotes the number of layers. Furthermore, due to the presence of non-linear activations in Adapter, this structure does not allow for re-parameterization, which leads to additional computational overhead in the inference. VPT~\cite{jia2022visual} incorporates \(m\) prompts into input space, leading to an increase of \(m\cdot D\) parameters for VPT-Shallow and \({m\cdot D\cdot L}\) for VPT-Deep. In contrast to Adapter, both LoRA~\cite{hu2021lora} and SSF~\cite{lian2022scaling} employ linear adaptation methods without incorporating non-linear functions. This design allows them to leverage re-parameterization benefits, thereby mitigating additional computations during inference. Specifically, the adaptation matrix of LoRA, which consists of a down-projection and an up-projection, introduces \({2\cdot w\cdot D\cdot D^\prime\cdot L}\) learnable parameters, where \(w\) denotes the number of attention matrices undergoing adaptation. SSF inserts linear scaling and shifting coefficients after \(o\) operations, resulting in an addition of \(2\cdot o\cdot D^*\cdot L\) extra parameters. \({D}^{*}\) denotes the dimension of weight matrix, where \({D}^{*} = 4\cdot D\) in up-projection of FFN and \({D}^{*} = D\) in other cases. 
ARC offers additional parameter compression by sharing symmetric projection matrices across different layers. This approach introduces \(D\cdot D^\prime\) parameters for MHA and FFN. The low-dimensional re-scaling coefficients and bias terms result in a total of \((D^\prime +D)\cdot L\) additional parameters. 
The proposed RLRR introduces dual-sided scaling tuning resulting in \(3\cdot o\cdot D^*\cdot L\) trainable parameters.

\section{Experimental details on larger-scale and hierarchical ViT backbones}
Table~\ref{tab:Ldetail}, ~\ref{tab:Hdetail}
and~\ref{tab:Swindetail} respectively display the comprehensive results of the comparison conducted in Section~\ref{sec:exp} among ViT-Large, ViT-Huge, and Swin-Base models.

\section{Expanded experiments with self-supervised pre-training}

In addition to the models pre-trained with supervision, we also conduct experiments with self-supervised pre-training approaches: MAE~\cite{he2022masked} and Moco V3~\cite{chen2021empirical}. 
Specifically, We utilize MAE and Moco V3 self-supervised pre-trained ViT-B as the backbone and evaluate the performance of our RLRR on VTAB-1k. The results of MAE and Moco V3 self-supervised models are presented in Table~\ref{tab:Maedetail} and Table~\ref{tab:Mocodetail}, respectively. 
Based on these results, it is noted that our proposed RLRR continues to exhibit competitive performance on two self-supervised ViT models.

\section{Flexibility of RLRR}
LoRA, as a universal fine-tuning paradigm, has achieved remarkable performance across multiple tasks due to its flexibility. 
In this section, we will elaborate on how our proposed RLRR maintains the flexibility comparable to LoRA while achieving superior performance. LoRA adjusts the trainable parameter count by altering the sampling dimensions of the bottleneck structure. Similarly, RLRR can achieve the same adjustment. Initially, we remove the $\mathbf{W}$ in fine-tuning items $\bigtriangleup\mathbf{W} = \vec{\boldsymbol{s}}_{\rm left}\odot\mathbf{W}\odot\vec{\boldsymbol{s}}_{\rm right}^{\top}$, defining this baseline as $\mathbf{X}(\mathbf{W}+ \mathbf{S}_{\rm left}\mathbf{S}_{\rm right})+\vec{\boldsymbol{b}}^{\top}+\vec{\boldsymbol{f}}^{\top}$ to simulate the LoRA rank = 1 scenario.

After this, we can also introduce variations in the RLRR variant by modifying the dimension $r$ of the parameter scaling in the expression $\mathbf{X}(\mathbf{W}+ (\mathbf{S}_{\rm left}\mathbf{S}_{\rm right})\odot\mathbf{W})+\vec{\boldsymbol{b}}^{\top}+\vec{\boldsymbol{f}}^{\top}$, where $\mathbf{S}_{\rm left}\in \mathbb{R}^{d \times r}$ and $\mathbf{S}_{\rm right}\in \mathbb{R}^{r \times d}$. Through this modification, we can derive adaptation matrices with varying ranks to demonstrate the flexibility of adjustments similar to LoRA. The results of above RLRR variants are shown in Table~\ref{tab:R1_lora}, which validate our statement.

\begin{table}[htbp] \scriptsize
  \centering
  \caption{Ablation study on VTAB-1k to compare with baseline.}
    \begin{tabular}{c|ccc|c}
    \toprule
    Method & Natural (7) & Specialized (4) & Structed (8) & Params \\
    \midrule
    w/o W & 81.3  & 85.5  & 57.5  & 0.33 \\
    w/ W  & \textbf{82.7} & \textbf{85.8} & \textbf{61.8} & 0.33 \\
    LoRA (r=16) & 80.4  & 85.2  & 61.0  & 0.63 \\
    \bottomrule
    \end{tabular}%
  \label{tab:R1_lora}%
\end{table}%

\begin{table}[htbp] 
  \centering
  \caption{Performance comparison of RLRR extended to CNNs.}
  \resizebox{\linewidth}{!}{
    \begin{tabular}{c|cc|cc}
    \toprule
     \multirow{2}[2]{*}{Methods} & \multicolumn{2}{c|}{ResNet-18} & \multicolumn{2}{c}{ResNet-50} \\
    & CIFAR-100 & Params & CIFAR-100 & Params \\
    \midrule
       Full fine-tuning & \textbf{79.7} & 11.23  & 80.7  & 23.71  \\
      Linear probing & 62.1  & 0.05  & 66.8  & 0.21  \\
      RLRR(r=1) & 75.0  & 0.08  & 79.0  & 0.27  \\
      RLRR(r=10) & 78.9  & 0.29  & \textbf{82.4} & 0.85  \\
    \bottomrule
    \end{tabular}%
    }
  \label{tab:rlrr_cnn}%
\end{table}%

\section{Transferability Analysis}
\begin{figure}[htbp]
\centering
\includegraphics[width=0.95\linewidth]{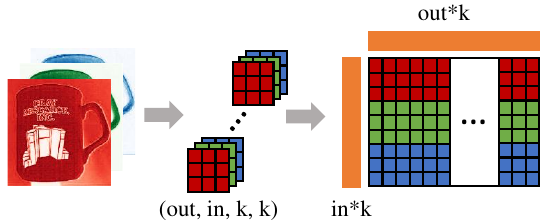}
\caption{Illustration of the RLRR method's extension to CNN.}
\label{rlrr_cnn_fig}
\end{figure}

We extend the RLRR variant to CNN by concatenating CNN kernels. 
As shown in Fig.~\ref{rlrr_cnn_fig}, by concatenating the convolutional kernels, we transform original convolutional kernel parameters in CNN to a two-dimensional parameter matrix $\mathbf{W'}$, allowing RLRR to seamlessly migrate to CNNs. Based on this, we supplement the experiments on CIFAR-100, as shown in Table~\ref{tab:rlrr_cnn}, which demonstrates the transferability of our RLRR on other deep learning model. We will explore applying our approach in future work under the field of NLP.
The outcomes of this variant are presented in Table~\ref{tab:rlrr_cnn}, underscoring the versatility of our design.

\section{Combination of multiple RLRRs}
RLRR can be likened to a LoRA with rank = 1. Consequently, operations like element-wise combination and arithmetic applied to LoRAs, as demonstrated in LCM-LoRA\cite{luo2023lcm}, LoRAHub \cite{huang2023lorahub}, and Composing PEMs\cite{zhang2024composing}, are also applicable to PLRR. The various combinations of RLRRs within their respective frameworks can be expressed in the form of
$\hat{\mathbf{S}}_{\rm left}\hat{\mathbf{S}}_{\rm right}=\sum_i^N w_i \mathbf{S}_{\rm left}^{i} \sum_i^N w_i \mathbf{S}_{\rm right}^{i}$ with $\hat{\vec{\boldsymbol{f}}}^{\top} = \sum_i^N w_i \vec{\boldsymbol{f}}_{i}^{\top}$, and $\hat{\mathbf{S}}_{\rm left}\hat{\mathbf{S}}_{\rm right}=\sum_i^N \mathbf{S}_{\rm left}^{i}\mathbf{S}_{\rm right}^{i}$ with $\hat{\vec{\boldsymbol{f}}}^{\top} = \sum_i^N w_i \vec{\boldsymbol{f}}_{i}^{\top}$.

\clearpage
\end{appendix}